\documentclass[runningheads]{llncs}
\usepackage{graphicx}

\usepackage{tikz}
\usepackage{comment}
\usepackage{amsmath,amssymb} 
\usepackage{color}
\usepackage{adjustbox}
\usepackage{float}

 \usepackage[width=122mm,left=12mm,paperwidth=146mm,height=193mm,top=12mm,paperheight=217mm]{geometry}

\usepackage{booktabs}
\usepackage{arydshln}
\usepackage{hyperref}
\usepackage[capitalise]{cleveref}
\usepackage{graphicx}
\usepackage{xspace}
\usepackage{todonotes}
\usepackage{mathtools}
\usepackage{tabularx}
\usepackage{multirow}
\usepackage{subfigure}

\begin{document}
\pagestyle{headings}
\mainmatter
\def\ECCVSubNumber{5160}  

\title{CoGS: Controllable Generation and Search from Sketch and Style}

\newcommand{\etal}{{\em et~al.}\xspace}
\newcommand{\eg}{e.\,g.\xspace}
\newcommand{\ie}{i.\,e.\xspace}


\titlerunning{CoGS: Controllable Generation and Search from Sketch and Style}

\author{Cusuh Ham\inst{1}\thanks{Equal contribution.} \and
Gemma {Canet Tarr\'es}\inst{2}$^\star$ \and
Tu  Bui\inst{2} \and
\\ 
James Hays\inst{1} \and
Zhe Lin\inst{3} \and
John Collomosse\inst{2,3}}
\authorrunning{Ham et al.}
%
\institute{Georgia Institute of Technology \email{\{cusuh,hays\}@gatech.edu} \and
University of Surrey \email{\{g.canettarres,t.v.bui,j.collomosse\}@surrey.ac.uk} \and
Adobe Inc. \email{zlin@adobe.com}}
\maketitle

\begin{abstract}
We present CoGS, a novel method for the style-conditioned, sketch-driven synthesis of images.  CoGS enables exploration of diverse appearance possibilities for a given sketched object, enabling decoupled control over the structure and the appearance of the output.  Coarse-grained control over object structure and appearance are enabled via an input sketch and an exemplar ``style'' conditioning image to a transformer-based sketch and style encoder to generate a discrete codebook representation.  We map the codebook representation into a metric space, enabling fine-grained control over selection and interpolation between multiple synthesis options before generating the image via a vector quantized GAN (VQGAN) decoder.  Our framework thereby unifies search and synthesis tasks, in that a sketch and style pair may be used to run an initial synthesis which may be refined via combination with similar results in a search corpus to produce an image more closely matching the user's intent.  We show that our model, trained on the 125 object classes of our newly created Pseudosketches dataset, is capable of producing a diverse gamut of semantic content and appearance styles.

\end{abstract}

\section{Introduction}

Generative artwork is a transforming creative practice, driven by advances in deep networks that enable diverse and realistic image synthesis \cite{Chen2018SketchyGAN,dalle,attngan,infogan,huang2021poegan,xue2022deep}. Sketches offer an intuitive modality to both control visual synthesis, and to search visual content \cite{Bui2018,Ribeiro2020Sketchformer,SongBMVC16}.  However sketches are inherently ambiguous, offering incomplete descriptions of users’ intent \cite{Collomosse2008,sketchstyle2017}.  While a sketch may communicate a rough approximation of structure or semantic layout, it offers limited fine-grained control over appearance or texture.  This presents a barrier to the practical use of sketches as a tool for controlled search and synthesis.  Thus, sketches are limited to serendipitous ``content discovery'' use cases, rather than as a tool to obtain an image with content and appearance matching a users' target intent. 

\begin{figure}[t!]
    \centering
    \includegraphics[width=\linewidth]{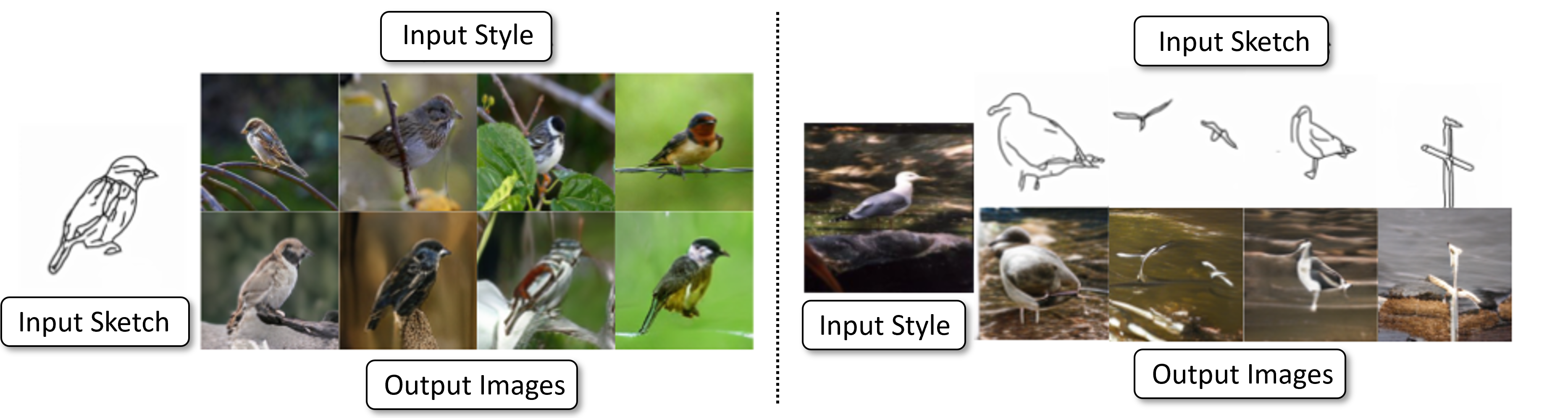}
    \caption{Images synthesized using CoGS. On the left we demonstrate the ability to control the style of the output for a given sketch, and on the right we demonstrate the ability to control the structure via multiple sketches for a given style image.}
    \label{fig:teaser}
\end{figure}

Inspired by the recently proposed DALL-E architecture \cite{dalle} for diverse, text-driven image synthesis, we introduce CoGS, a novel method for diverse, sketch-driven image synthesis with fine-grained control over structure and appearance (Fig.~\ref{fig:teaser}).  We make the following technical contributions:

1. Style-conditioned sketch-based synthesis using Vector Quantized GANs (VQGANs) \cite{esser2020taming}.  We synthesize images via a VQGAN decoder, driven by discrete codebook representations generated via a transformer-based sketch and style encoder. We show decoupled coarse-grained control over the structure and appearance of the synthesized image using a provided sketch and exemplar ``style'' image.  This enables users to explore multiple appearance possibilities for a given sketched object, and to do so for a diverse set of object classes.

2. Unified embedding for search and synthesis.  We propose a variational auto-encoder (VAE) \cite{VAE} module that maps codebook representations into a metric space. The embedding thus enables fine-tuning the appearance of the synthesized image either by exploring the local space, or by interpolating between existing similar images within a search corpus.

3. Paired sketch-image dataset. To enable training and evaluation of our model, we create a novel paired dataset of images, hand-drawn sketches, and ``pseudosketches'' derived from the images via an automated process and graded via crowdsourcing. Specifically, the dataset comprises of 113,370 images and corresponding pseudosketches.  A subset of 9,580 images map to the existing Sketchy Database \cite{sangkloy2016}, providing 5-10 free-hand sketches for each image.  

\section{Related Work}

{\bf Sketch-based image generation} was first explored through non-parametric patch-based approaches initially developed for texture synthesis \cite{Efros2001,Wexler_CVPR_2004,Barnes_SIGGRAPH_2009,Hays_SIGGRAPH_2007}, and extended to image synthesis via visual analogy \cite{Hertzmann2001} and interactive montage \cite{chen2009sketch2photo}.  These methods recall and blend textures sampled from a training set, guided by labeled sketches or semantic maps \cite{patchsurvey}. With the advent of deep learning, image translation networks, such as CycleGAN \cite{CycleGAN2017} and pix2pix \cite{pix2pix2016,wang2018pix2pixHD}, were exploited to map sketches directly to photos for a single class \cite{xian2017texturegan,Sangkloy2017scribbler,ghosh2019isketchnfill}.  These methods were later adapted for high-resolution portrait synthesis (e.g., via convolutional inversion \cite{sketchinv} and semantic priors \cite{yangface}).  SketchyGAN \cite{Chen2018SketchyGAN} and ContextualGAN \cite{Lu2018ContextualGAN} proposed multi-class extensions for creating low-resolution outputs of individual objects, building upon the success of conditional generative adversarial networks (cGANs) \cite{Mirza2014}.  Zhu \etal proposed a manifold exploration technique to improve controllability of cGAN synthesis \cite{Zhu2016control}.  Gao \etal showed that these approaches were insufficient for compositing scenes using objects of differing classes, proposing to fuse scene graph representations with cGANs for direct sketch-to-scene image synthesis \cite{Gao2020SketchyCOCO}.  Scene compositions were also produced by mapping sketches to semantic segmentation maps \cite{SceneDesignerSHE2021} using SPADE \cite{park2019SPADE}. 

{\bf Conditional image generation} has been explored for multiple input modalities.  Text embeddings \cite{Reed2016a,Reed2016b} have been incorporated into both generator and discriminator of cGANs for keyword-driven synthesis, and extended to natural language phrases.  A two-stage generator guided via attention from input features was proposed in \cite{attngan}. Images have been used for conditioning upsampling \cite{Gao2020} and inpainting \cite{Guo2021}, while sketches mostly focus on image colorization \cite{Iizuka2017}. The use of bounding box layouts \cite{layout1,layout2,layout3} and scene graphs \cite{sg1,sg2} has also been explored.
Recently, transformers have been explored for image translation and completion tasks, and combined with the discrete codebook representations of VQGAN \cite{esser2020taming}.  DALL-E \cite{dalle} exploited VQGAN to learn a direct mapping from natural language to image for diverse image synthesis.  Our work builds upon a similar concept, exchanging text for sketches and further conditioning the codebook generation on an appearance (style).  In this sense, our approach is aligned to recent instance-conditioned (IC) synthesis.  IC-GAN \cite{icgan} performed retrieval using descriptors derived from the source image, using an average of spatial semantic maps derived from the top results.  Our method optionally leverages retrieval to interpolate between images similar to the synthesized output to fine-tune appearance. Recently, PoE-GAN \cite{huang2021poegan} has also explored multiple guidance cues (sketch, semantic map, text) to synthesize images.  Our approach differs both in our transformer architecture and in our two-stage control, offering coarse-grained conditioning on synthesis using a sketch and style image and using a continuous embedding for fine-grained refinement.

\section{Methodology}

CoGS accepts a labelled sketch (a raster, and an associated semantic class label), and a style image as inputs. Given these three conditioning signals, our goal is to synthesize an image of the specified class that combines the general structure of the sketch with the colors and textures of the style image, providing users with decoupled control over both the content and style of the output image. An optional additional step can be used to further refine the output.

\begin{figure}[t!]
    \centering
    \includegraphics[width=\linewidth,clip,height=7cm]{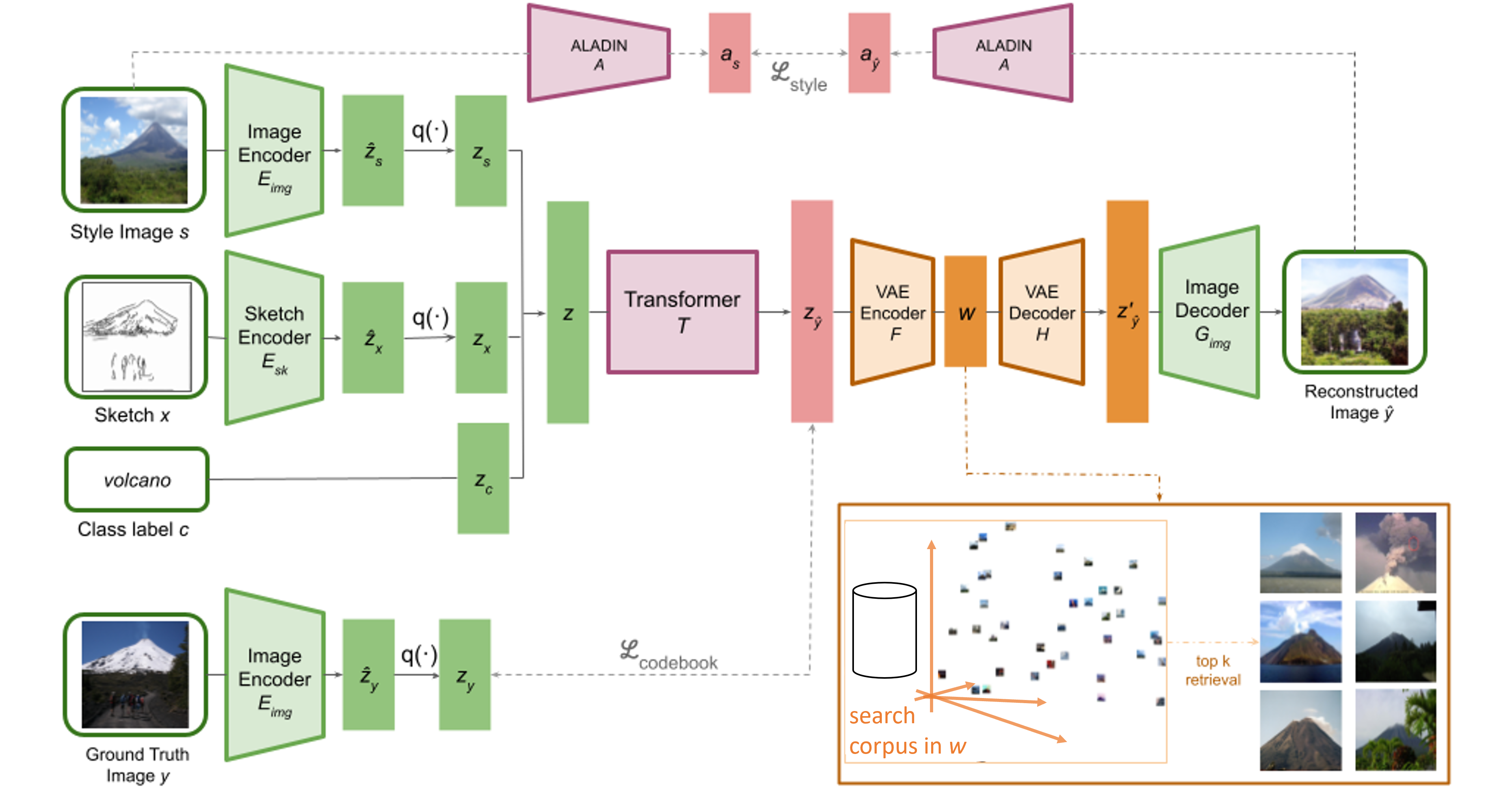}
    \caption{Illustration of the full CoGS pipeline. We use VQGANs (shown in green) to learn two codebooks, one for sketches and one for images, which are combined with a tokenized representation of a class label. Then, we use a transformer with an auxiliary style loss (shown in red) to learn to composite the inputs into a predicted codebook, which can be subsequently decoded by the image VQGAN decoder to synthesize an image. The input structure and style images offer coarse-grained control over the synthesis.  Finally, we use a VAE (shown in orange) to map the codebook to and from a latent space that enables fine-grained refinement of the output image via retrieval or interpolation of results from a search corpus.}
    \label{fig:pipeline_diagram}
\end{figure}

A major challenge for this task is understanding the correspondence between the input sketch and style image. That is, the network must understand how to appropriately map textures from the style image to the corresponding regions of the sketch. We propose to feed an additional input of a class label to resolve correspondence ambiguities, and to clarify the object of interest. CoGS consists of three components (\cref{fig:pipeline_diagram}): 1) VQGANs for encoding the sketch and style inputs; 2) a transformer network for sketch- and style-conditioned image synthesis; and 3) a VAE to further refine output through retrieval and interpolation.

\subsection{Paired Pseudosketches dataset}

While several datasets of human-drawn sketches exist, they are often too small, span only a single or small number of categories \cite{sketchshoe,Song_2017_ICCV}, contain simplistic stock images \cite{Song_2017_ICCV,sketchshoe,ghosh2019isketchnfill}, or lack paired image correspondences \cite{googlequickdraw,eitz2012hdhso}. For our work, we require a large-scale dataset of sketch-image pairs across diverse categories in order to learn to synthesize images that capture the structural characteristics of the input sketch. The most relevant existing dataset is the Sketchy Database \cite{sketchydb}, which contains over 75K sketches of 12.5K images across 125 categories.

\begin{figure}[t!]
    \centering
    \includegraphics[width=\linewidth,height=4cm]{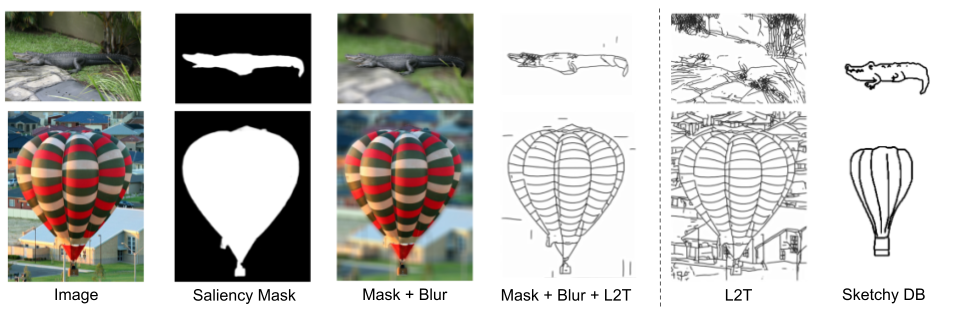}
    \caption{Stages in the creation of the Pseudosketches dataset. Given an input image, we generate its saliency mask with \cite{selectsubject}, blur the non-salient regions, and extract the edgemap of the masked image using L2T \cite{sumie}. Without the saliency mask blurring, we would retain background details not present in hand-drawn sketches.}
    \label{fig:dataset}
\end{figure}

The limited number of images in Sketchy DB makes it difficult for a network to learn to synthesize a diverse set of photorealistic outputs. Thus, we propose to create a new dataset of sketch-like edgemaps, or ``pseudosketches'', extracted from a larger pool of images to create a large set of explicit sketch-image pairs. We take the 12.5K images from Sketchy DB and query additional images from similar ImageNet categories (based on WordNet distances) to run through our automated extraction pipeline shown in \cref{fig:dataset}.

For each image, we generate a saliency mask using an example-based open-set panoptic segmentation network \cite{selectsubject}, and blur the non-salient regions with a large Gaussian kern. Then, we run a line drawing extractor \cite{sumie} to create pseudosketches of the masked images. Blurring is an essential step in making the pseudosketches more similar in nature to hand-drawn sketches. Without blurring, we retain extraneous background details as shown in \cref{fig:dataset}. Finally, we present the input image and corresponding pseudosketch side-by-side to Amazon Mechanical Turk (AMT) workers, and ask them to rate how representative the pseudosketch is of the image on a scale of 1 (lowest) to 5 (highest), given the semantic class of the image. After filtering out any pseudosketches with a score below 3, we have 113,370 pseudosketch-image pairs across the original 125 categories of Sketchy DB.

\subsection{Learning effective encodings for input modalities} \label{sec:vqgan}

Given a sketch $x$, style image $s$, and class label $c$ as inputs, we need to encode and combine each modality into a single input to the network that synthesizes a composite image. Inspired by recent success with visual sequential modeling \cite{dosovitskiy2020image,dalle}, we want to represent the inputs as a sequence of tokens. We use two vector quantized generative adversarial networks (VQGANs), one trained on pseudosketches and one trained on images, to encode the sketch and style inputs, respectively, into codebook representations. The class label is encoded as a single token representing the index of the class.

Each VQGAN consists of an encoder $E$, decoder $G$, and discriminator $D$. The goal of $E$ and $G$ is to learn the best way to represent an image $x \in \mathbb{R}^{H \times W \times 3}$ as a collection of codebook entries $z_x \in \mathbb{R}^{h \times w \times n_z}$, where $n_z$ is the dimensionality of codes. The discriminator $D$ is trained to ensure that the learned codebook $\mathcal{Z} = {\{ z_k\}}^K_{k=1} \subset \mathbb{R}^{n_z}$ is both as rich and compressed as possible, encouraging high quality generative outputs through adversarial training with $G$.

A quantization step $q(\cdot)$ converts the continuous output of the encoder $\hat{z}_x = E(x) \in \mathbb{R}^{h \times w \times n_z}$ into its discretized form $z_x = q(\hat{z}_x)$ by considering the closest codebook entry $z_k$ to each spatial code of $\hat{z}_x$. Then, $G$ can be trained so that we obtain a reconstruction $\hat{x} = G(z_x)$ as close as possible to the original image $x$. An additional patch-based discriminator $D$ and perceptual loss are applied to $\hat{x}$ to further ensure that the codebook entries capture details that contribute to improving the overall quality and realism of the image.

\subsection{Synthesizing images with transformers} \label{sec:transformer}

In order to train the CoGS transformer, we must select an image that represents the style connecting each sketch to its corresponding image. We use a pre-trained ALADIN \cite{Ruta2021ALADINAL} style encoder $A$ to pre-compute the style embedding $a_y = A(y)$ for all images $y$ in the Pseudosketches dataset. Then we compute the pairwise Euclidean distance between all style embeddings $\{a_y^c\}$ belonging to the same class $c$ to find the nearest neighbor $s$ for each $y$. By limiting the nearest neighbor to come from within the same class, $s$ is guaranteed to be semantically and stylistically similar to $y$ but not necessarily share the same fine-grained details of the object (e.g., pose, orientation, scale). Thus, the network must learn a more complex relationship between the sketch and style inputs to apply the textures of the style image to the corresponding regions of the sketch.

After determining an appropriate style image for each sketch-image pair, we now have inputs of a sketch $x$, style image $s$, and class label $c$ whose combined conditioning signals should produce a target image $y$. We freeze the weights of the VQGANs described in \cref{sec:vqgan}, and use the encoders $E_{sk}$ and $E_{img}$ to encode the sketch and style images into $\hat{z}_x$ and $\hat{z}_s$, respectively. The quantization step $q(\cdot)$ discretizes the representations into sequences of indices from the codebook $z_x = q(\hat{z}_x)$ and $z_s = q(\hat{z}_s)$ by replacing each code by its index in their respective codebooks, $\mathcal{Z}_{sk}$ and $\mathcal{Z}_{img}$, before concatenating them with the tokenized class encoding $z_c$ into a single input $z$ to a transformer $T$.

Given a concatenated input $z$ for some input $(x, s, c)$, the transformer $T$ aims to learn $z_{y} = E_{img}(y)$, which is the codebook representation of the image $y$ corresponding to the input pseudosketch $x$. The likelihood of the target sequence $z_{y}$ is modeled autoregressively as

\begin{equation}
    p(z_y|x, s, c) = \prod_i p(z_y^i | z_y^{<i}, x, s, c).
\end{equation}

Thus, the codebook loss is defined as maximizing the log-likelihood:

\begin{equation}
    \mathcal{L}_{\text{codebook}} = \mathbb{E}_{y\sim p(y)}[-\log p(z_y | x, s, c)].
    \label{eq:codebook}
\end{equation}

We also introduce an auxiliary style loss $\mathcal{L}_{style}$ to further reinforce the style constraint on the generated image. We decode the predicted codebook $z_{\hat{y}} = T(\{z_x, z_s, z_c\})$ using the image VQGAN decoder $G_{img}$ into image $\hat{y} = G_{img}(z_{\hat{y}})$, and compute its ALADIN style embedding $a_{\hat{y}} = A(\hat{y})$ as well as the ALADIN embedding of the input style image $a_{s} = A(s)$. Then, the style loss is defined as:

\begin{equation}
    \mathcal{L}_{\text{style}} = \text{MSE}(a_{s}, a_{\hat{y}}),
    \label{eq:style}
\end{equation}

where $\text{MSE}(i, j)$ is the mean squared error between $i$ and $j$. Thus, the total transformer loss is defined as the sum of the two losses weighted by $\lambda_T$:

\begin{equation}
    \mathcal{L}_{\text{transformer}} = \mathcal{L}_{\text{codebook}} + \lambda_T\mathcal{L}_{\text{style}}.
    \label{eq:transformer}
\end{equation}

\subsection{Refining outputs with VAEs} \label{sec:vae}

The decoded image from the transformer's output $\hat{y} = G_{img}(z_{\hat{y}})$ may not always lead to the exact result in mind, whether a consequence of the difficulty of the task or the user wanting to make additional modifications. Therefore, we provide an optional step to further refine the generated image via retrieval or synthesis.

Freezing the VQGANs and transformer from the earlier stages of the pipeline, we train a variational autoencoder (VAE) \cite{VAE} with encoder $F$ and decoder $H$ to map the encoding $z_{\hat{y}}$ to a more compact representation $w = F(z_{\hat{y}}) \in \mathbb{R}^{d}$.
The VAE is trained on synthesis and search to ensure $w$, corresponding to the latent space $\mathcal{W} \subset \mathbb{R}^{d}$, constitutes a unified embedding for both tasks.

The VAE encoder output $w$ is a distribution represented by two vectors: the mean $w_{\mu} \in \mathbb{R}^{d}$, and standard deviation $w_{\sigma} \in \mathbb{R}^{d}$. The traditional evidence lower bound (ELBO) objective of VAEs, defined in \cref{eq:ELBOloss}, encourages a smooth latent space $\mathcal{W}$ that can both reconstruct an input image and synthesize novel outputs:

\begin{equation}
    \mathcal{L}_{\text{ELBO}} = \mathcal{L}_\text{KL} - \mathcal{L}_{\text{rec}} = \min \mathbb{E}_p [\log{p(w | z_{\hat{y}}^q)} - \log{g(w)}] - \mathbb{E}_q \log{g(z_{\hat{y}}^q|w)}.
    \label{eq:ELBOloss}
\end{equation}

While $\mathcal{L}_{ELBO}$ enables synthesis, we must also enforce metric properties on $\mathcal{W}$ to enable it for retrieval. We leverage self-supervised contrastive learning \cite{pmlr-v119-chen20j,isola2019,henaff2019dataefficient,hjelm2018learning} to map embeddings of structurally similar images close together in latent space and embeddings of dissimilar images far apart:

\begin{equation}
   \mathcal{L}_{\text{contrastive}} = - \sum_{i \in I} \log{\frac{\exp{(w_i \cdot w_{j(i)}/\tau)}}{\sum_{a \in I \backslash {i} }\exp{(w_i \cdot w_a/\tau)}}}.
   \label{eq:selfsupervisedcontrastive}
\end{equation}

For an anchor $i$ in a batch of $2N$ elements, a positive $j(i)$ is an image generated by the CoGS transformer using the same sketch and different style images as $i$, while the remaining $2(N-1)$ elements are negatives generated using different sketches and completely different style images. Thus, the VAE loss is defined as the sum of the two losses weighted by $\lambda_V$:

\begin{equation}
    \mathcal{L}_{\text{VAE}} = \mathcal{L}_{\text{ELBO}} + \lambda_V \mathcal{L}_{\text{contrastive}}.
    \label{eq:VAEloss}
\end{equation}

\section{Experiments}

We describe our experimental setup, evaluation protocols, and comparisons to baseline methods. We evaluate the quality and ability of the transformer component of CoGS to enable both style and structure controllability in \cref{sec:exp-partition}-\ref{sec:exp-ablation}, and investigate the VAE refinement step in \cref{sec:exp-refinement}.

\subsection{Experimental setup}

\textbf{Network architectures and training.} We use an ImageNet pre-trained VQGAN from \cite{esser2020taming} for $E_{img}$ and $G_{img}$, and train another VQGAN on the Pseudosketches dataset for $E_{sk}$. Both VQGANs are trained with the same settings: the encoder $E$ transforms $x \in \mathbb{R}^{256 \times 256 \times 3}$ to codebooks $z_x \in \mathbb{R}^{16 \times 16 \times 256}$, and the decoder $G$ reconstructs $z_x$ into $\hat{x} \in \mathbb{R}^{256 \times 256 \times 3}$. For the transformer, we use 16 layers with 16 attention heads, and a vocabulary size $|\mathcal{Z}| = 1024$. The dimensionality of the sketch and style codebooks is $n_{\hat{z}_x} = n_{\hat{z}_s} = 256$ and class token $n_{z_c} = 1$, so the concatenated sequences of quantized inputs are of length $n_z = 513$. We randomly partition the Pseudosketches dataset into 102,024 training and 11,346 validation examples, and use a weighting term $\lambda_T=1$ in \cref{eq:transformer}.

The VAE encoder $F$ compresses input codebooks $z_y \in \mathbb{R}^{16 \times 16 \times 256}$ into $w \in \mathbb{R}^{1024}$, and the decoder $H$ reconstructs the input back to $z'_y \in \mathbb{R}^{16 \times 16 \times 256}$. With the three components of CoGS being trained sequentially, the VAE is trained using a two-stage approach where training is bootstrapped by only using the contrastive loss $\mathcal{L}_{contrastive}$ for training the encoder $F$. When the loss converges, both $F$ and $H$ are further trained using the dual loss in \cref{eq:VAEloss} with $\lambda_V=10^6$.

\textbf{Evaluation metrics.} Two popular metrics to assess the quality of generative networks are the Fr\'echet Inception Distance (FID) \cite{heusel2017gans} and Learned Perceptual Image Patch Similarity (LPIPS) \cite{zhang2018unreasonable}. FID computes the distance between the distribution of synthesized validation images and the distribution of the ground truth images corresponding to the input sketches. Thus, the lower the FID, the more similar the generated images are to the distribution of real images.

LPIPS measures the distance between image patches, and is used to quantify a generative network's ability to produce diverse outputs for a set of inputs. We subsample the top 10\% validation pseudosketches (as determined by AMT scores during data collection) and their respective style images and class labels, sample 5 output images per input, and average the LPIPS on all unique pairs of outputs.

We further evaluate CoGS using style and structure distance metrics (described in \cref{sec:exp-controllability}), and AMT to crowd-source quality assessments of CoGS and the baselines.  For the latter we source both: 1) user preference for our approach versus baseline approaches, and 2) a subjective evaluation methodology akin to Gao \etal \cite{Gao2020SketchyCOCO} that scores how ``realistic'' each image is, and how ``faithful'' each synthesized image is to its conditional inputs and ground truth image.

\subsection{Dataset partitioning} \label{sec:exp-partition}

To approximate the difficulty for the CoGS transformer to synthesize the various categories of the Pseudosketches dataset, we compute the FID on the validation set for each of the 125 classes individually, and group them into 3 sets: the ``simple'' partition contains classes with the 41 lowest FIDs, ``medium'' with the 42 middle FIDs, ``complex'' with the 42 highest FIDs. We then combine the images from all the classes within each partition and recompute the overall FID score of each partition, shown in \cref{tab:baselinesquant}. We note that there is a correlation between the number of images in a class and its FID score, i.e., classes with higher number of images tend to produce lower FIDs.

\begin{figure}[t!]
    \centering
    \includegraphics[width=\linewidth,height=7.5cm]{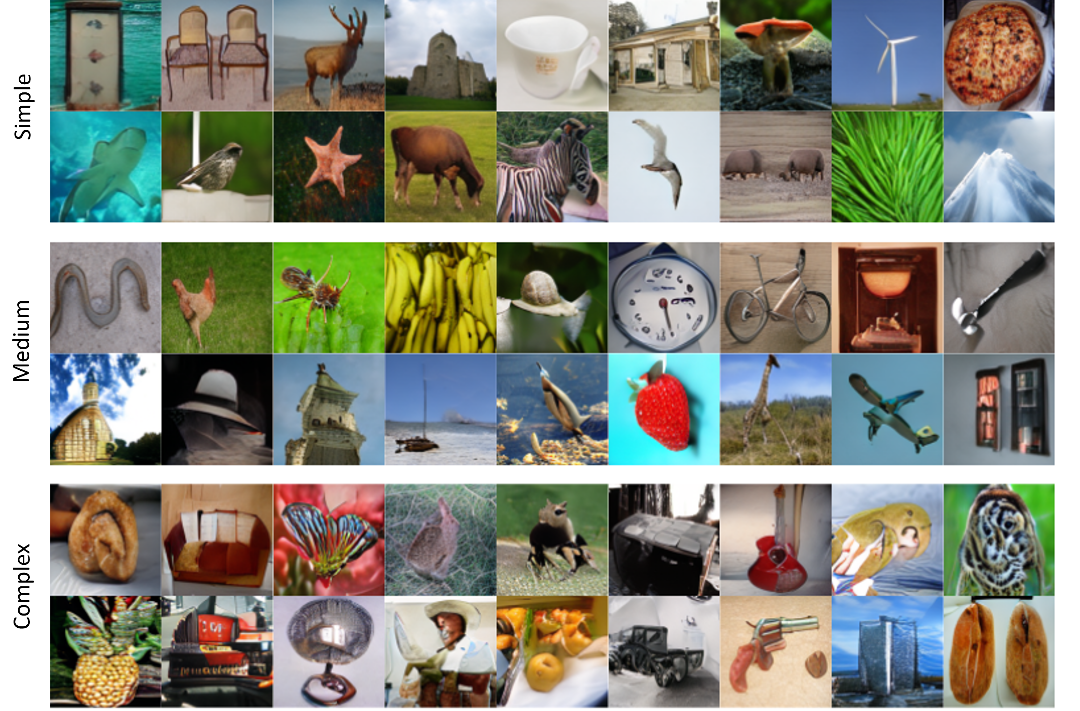}
    \caption{Representative examples generated by the CoGS transformer stage for the simple, medium, and complex partitions.}
    \label{fig:partitions}
\end{figure}

\subsection{Comparison to baseline methods}

We briefly describe the baseline methods used for evaluations. Other methods such as PoE-GAN \cite{huang2021poegan} and Scribbler \cite{Sangkloy2017scribbler} align with our work, but lack public code or models for comparison.

\textbf{Neural Style Transfer} (NST) \cite{GatysNST} takes as input a content image $x_c$ and style image $x_s$, and synthesizes an image $\hat{x}$ where the style of $x_s$ is ``transferred'' onto $x_c$ by optimizing it to match texture (Gram matrix) statistics from $x_s$ using mid-late layer activations of VGG-16 \cite{simonyan2014very}. We use the pseudosketch as the input.

\textbf{SketchyGAN} \cite{Chen2018SketchyGAN} is a sketch-conditioned GAN trained on Sketchy DB and edgemaps of Flickr images. We substitute Flickr edgemaps for the Pseudosketches dataset to train SketchyGAN as the data and model are not public.

\textbf{iSketchNFill} \cite{ghosh2019isketchnfill} is a method for inpainting partial sketches and synthesizing images from them. We train the generative component on the Pseudosketches dataset, though the wide class diversity proved challenging for this network.

\textbf{Instance-Conditioned GAN} (IC-GAN) \cite{icgan} is a conditional GAN that leverages instances of images or text to condition the generated output, guided by pre-computed descriptors from the source images and their top retrieval results. We use the Pseudosketches dataset to train two variants that still compute the descriptors on photorealistic images: 1) add the corresponding pseudosketch's descriptors as an additional conditioning instance, and 2) replace the main photorealistic input with its corresponding pseudosketch.

\textbf{CoCosNet v2} \cite{cocosnetv2} is a patch-based method that uses examplar images from different domains, such as edge maps or semantic maps, to translate into high-quality photorealistic images.

\begin{figure}[t!]
    \centering
    \includegraphics[width=\linewidth,height=6.75cm]{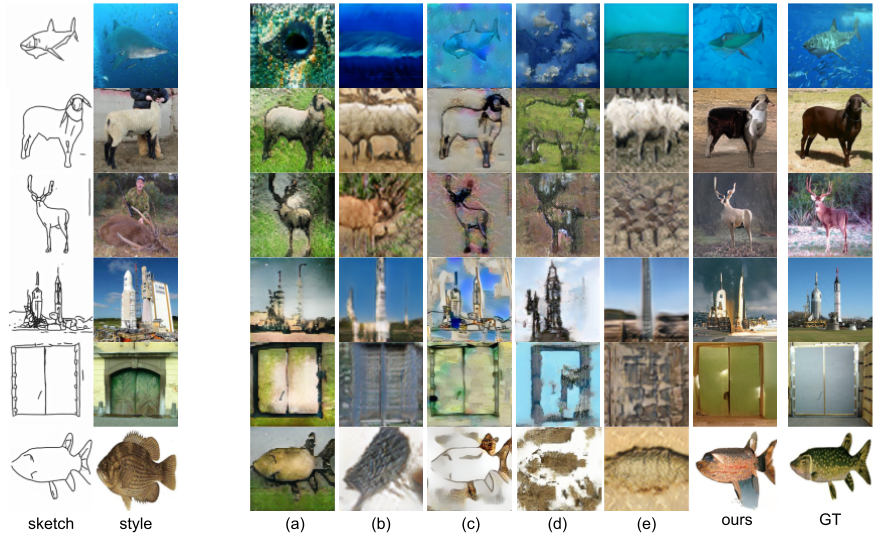}
    \caption{Representative outputs from: a) SketchyGAN \cite{Chen2018SketchyGAN}, b) IC-GAN \cite{icgan} (sketch), c) NST \cite{GatysNST}, d) CoCosNetv2 \cite{cocosnetv2}, e) IC-GAN \cite{icgan} (sketch, style image), and CoGS. Note that neither (a) or (b) use the input style image. We emphasize that the synthesized image may not necessarily aim to look exactly like the ``ground truth'' image due the stylistic differences between the style and ground truth images (e.g., second row style image contains a white sheep, but the ground truth image is a black sheep).}
    \label{fig:baselines}
\end{figure}

Quantitative evaluations using FID and LPIPS are provided in \cref{tab:baselinesquant}, and qualitative evaluations based on AMT experiments in \cref{tab:baselinesqual}. Our method produces the best FID and comparable LPIPS, indicating good overall quality of images while avoiding mode collapse by producing diverse outputs. The slightly higher LPIPS score from SketchyGAN is likely due to the output being constrained only by the sketch input, whereas the solution space of CoGS is reduced by imposing additional style and class conditions. 

We ran three crowd-sourced AMT evaluations: 1) \textit{Realism}: We present output images (\eg \cref{fig:baselines}) to participants, and ask them to rate the realism  on a scale 1 (``very dissatisfied'') to 5 (``very satisfied'') using 3 participants per image.  We consider only responses with consensus, where the maximum and minimum ratings deviate by at most 1 point; 2) \textit{Fidelity}: We similarly measure the fidelity, or ``faithfulness'', of the output to the ground truth image.
Although scores are generally low (2-3 on the scale), they are similar to scores reported in other papers following this methodology (e.g., \cite{SceneDesignerSHE2021,Gao2020SketchyCOCO}); and 3) \textit{Preference}: We ask 3 participants per image to indicate their preferred method in terms of overall fidelity to the ground truth image (gt), fidelity of structure to the sketch (sk) and to the style (st).  Responses with majority consensus (2 out of 3 agree) are included.
CoGS outperforms the baselines in the majority of the AMT studies, and produces the highest ratings overall. NST achieves higher ratings on faithfulness and SketchyGAN on structure only on the complex set.

\begin{table}[t!]

\centering
\begin{adjustbox}{width=0.96\textwidth}

\begin{tabular}{lcccccccc}
\toprule
\multicolumn{1}{l}{}  & \multicolumn{2}{c}{\textbf{Simple }}  & \multicolumn{2}{c}{\textbf{ Medium}}  &  \multicolumn{2}{c}{\textbf{ Complex}}  & \multicolumn{2}{c}{\textbf{ Overall}}  \\ 
\cmidrule{2-9}
\textbf{Method}  &  \textbf{FID$\downarrow$}  &  \textbf{LPIPS$\uparrow$}  &  \textbf{FID$\downarrow$}  &  \textbf{LPIPS$\uparrow$}  &  \textbf{FID$\downarrow$}  &  \textbf{LPIPS$\uparrow$}  &  \textbf{FID$\downarrow$}  &  \textbf{LPIPS$\uparrow$}  \\ 
\cmidrule{1-9}
SketchyGAN $\circ$  &  210.919  &  \textbf{0.534}  &  256.167  &  \textbf{0.555}  &  330.649  &  0.550  &  213.762  &  \textbf{0.541}  \\ 
\cmidrule{1-9}
IC-GAN $\diamond$  & 103.106  &  0.183  & 161.924  &  0.209  & 293.037  &  0.185  &  109.336   &  0.190  \\
iSketchNFill $\diamond$  &  497.693  &  1e-8   &  522.380  &  9e-9  & 548.525  &  1e-8  &  506.790  &  1e-8  \\ 
\cmidrule{1-9}
NST $\bullet$  &  118.839  &  0.308  &  167.651  &  0.302  &  262.893  &  0.311  &  114.707  &  0.307  \\ 
CoCosNetv2 $\bullet$   &  153.371  &  0.399  &  204.983  &  0.404  &  279.311   &  0.403  &  160.705   &  0.401  \\ 
\cmidrule{1-9}
IC-GAN $\star$  &  131.750  &  0.180  &  176.123  &  0.216  &  293.086  &  0.189  &   130.235  &  0.190  \\
CoGS (ours) $\star$  &  \textbf{43.896}  &  0.500  &  \textbf{95.539}  &  0.547  &  \textbf{201.230}  &  \textbf{0.616}  &  \textbf{50.630}  &  0.521  \\
\bottomrule
\end{tabular}
\end{adjustbox}
\caption{Quantitative comparison of FID and LPIPS scores. The symbol next to each method denotes the input(s): $\circ = \{x\}$, $\diamond = \{x, c\}$, $\bullet = \{x, s\}$, and $\star = \{x, s, c\}$, where $x$ is a sketch, $s$ is a style image, and $c$ is the class label.}
\label{tab:baselinesquant}
\end{table}

\begin{table}[t!]
\begin{adjustbox}{width=1.0\textwidth}
\begin{tabular}{lccccccccccccccc} 
\toprule
\multicolumn{1}{c}{\begin{tabular}[c]{@{}c@{}}\\\end{tabular}} & \multicolumn{5}{c}{\textbf{Simple}}  &  \multicolumn{5}{c}{\textbf{Medium}}  &  \multicolumn{5}{c}{\textbf{Complex}}  \\ 
\cmidrule{2-16}
\textbf{Method}  &  \textbf{F}  &  \textbf{R}  &  \begin{tabular}[c]{@{}l@{}} \textbf{pref. } \\ \textbf{ (gt)}\end{tabular}  &  \begin{tabular}[c]{@{}l@{}} \textbf{pref. } \\ \textbf{ (sk)} \end{tabular}  &  \begin{tabular}[c]{@{}l@{}} \textbf{pref. } \\ \textbf{ (st)} \end{tabular}  &  \textbf{F}  &  \textbf{R}  &  \begin{tabular}[c]{@{}l@{}} \textbf{pref. } \\ \textbf{ (gt)} \end{tabular}  &  \begin{tabular}[c]{@{}l@{}} \textbf{pref. } \\ \textbf{ (sk)} \end{tabular}  &  \begin{tabular}[c]{@{}l@{}} \textbf{pref. } \\ \textbf{ (st)} \end{tabular}  &  \textbf{F}  &  \textbf{R}  &  \begin{tabular}[c]{@{}l@{}} \textbf{pref. } \\ \textbf{ (gt)} \end{tabular}  &  \begin{tabular}[c]{@{}l@{}} \textbf{pref. } \\ \textbf{ (sk)} \end{tabular}  &  \begin{tabular}[c]{@{}l@{}} \textbf{pref. } \\ \textbf{ (st)} \end{tabular}  \\ 
\cmidrule{1-16}
SketchyGAN $\circ$  &  2.68  &  2.67  &  23.37  &  36.06  &  7.04  &  2.52  &  2.87  &  31.72  &  38.32  & 10.84  &  2.64  &  2.83  &  36.21  &  \textbf{49.89}  &  16.19  \\ 
\cmidrule{1-16}
IC-GAN $\diamond$  &  2.49  &  2.52  &  5.66  &  0.56  &  22.26  &  2.27  &  2.63  &  6.04  &  0.82  &  21.76  &  2.36  &  2.62  &  3.60  &  0.67  &  16.97  \\ 
iSketchNFill $\diamond$  &  1.96  &  1.95  &  0.00  &  0.00  &  0.00  &  1.87  &  2.13  &  0.18  &  0.74  &  0.25  &  1.65  &  1.99  &  0.00  &  0.00  &  0.26  \\ 
\cmidrule{1-16}
NST $\bullet$  &  2.69  &  2.67  &  10.53  &  10.73  &  21.13  &  2.52  &  2.85  &  8.35  &  7.71  &  21.76  &  \textbf{2.68}  &  2.82  &  10.07  &  9.89  &  23.39  \\ 
\cmidrule{1-16}
IC-GAN $\star$  &  2.34  &  2.39  &  4.32  &  0.84  &  15.60  &  2.14  &  2.46  &  4.21  &  0.67  &  11.60  &  2.26  &  2.51  &  5.28  &  0.9  &  10.80  \\ 
CoGS (ours) $\star$  &  \textbf{2.94}  &  \textbf{3.04}  &  \textbf{55.93}  &  \textbf{51.80}  &  \textbf{33.96}  &  \textbf{2.67}  &  \textbf{3.05}  &  \textbf{49.44}  &  \textbf{52.40}  &  \textbf{33.78}   &  2.58  &  \textbf{2.92}  &  \textbf{44.60}  &  38.65  &  \textbf{32.39}  \\
\bottomrule
\end{tabular}
\end{adjustbox}
\caption{Qualitative evaluations within each partition and across the overall validation set based on AMT experiments. The considered evaluation metrics are: fidelity to ground truth (F), realism of the generated image (R), preference overall given the ground truth image (pref. (gt)), preference based on structure, given the input sketch (pref.(sk)), preference based on style, given the style image (pref.(st)). Values for pref. (gt)/(sk)/(st) correspond to percentages of workers' selection as the best option, while F and R correspond to workers' rankings of outputs from 1 (worst) to 5 (best).}
\label{tab:baselinesqual}

\end{table}

\subsection{Style and structure controllability} \label{sec:exp-controllability}

While the realism of the generated images is important, the main goal of the CoGS transformer is to provide users decoupled control over both the style and structure of the generated image across a diverse set of categories.

\begin{table}[t]
\centering
\begin{tabular}{ccccc}
\toprule
\textbf{Partition}  &  \textbf{Style}$\downarrow$  &  \textbf{AMT Style}$\uparrow$  &  \textbf{Structure}$\downarrow$  &  \textbf{AMT Structure}$\uparrow$ \\
\cmidrule{1-5}
Simple  &  1.085  &  2.655  &  2.627  &  3.117  \\
Medium  &  1.136  & 2.374  &  2.004  &  2.805  \\
Complex  &  1.104 &  2.450  &  1.519  &  2.393  \\
\cmidrule{1-5}
Overall  &  1.100  & 2.501&  2.387  &  2.776\\
\bottomrule
\end{tabular}
\caption{Evaluations for style and structure controllability using distance metrics (ALADIN and Chamfer distances, respectively) and AMT human evaluations where participants are asked to score the fidelity of the output to the input style and sketch on a scale of 1 (low) to 5 (high).}
\label{tab:controllability}
\end{table}

\textbf{Style controllability.} For measuring the stylistic similarity of the generated image to the input style image, we use $d_{style}(s, \hat{y}) = d(a_s, a_{\hat{y}})$, where $a_s$ and $a_{\hat{y}}$ are the style encodings from ALADIN \cite{Ruta2021ALADINAL} of the style and synthesized images, respectively, and $d(i, j)$ is the Euclidean distance between $i$ and $j$. In \cref{tab:controllability} we show that the mean style distance of each partition are all within a small epsilon of each other, which agrees with AMT style similarity evaluations, where workers are asked to rate how well the generated image matches the style image on a scale of 1 (worst) to 5 (best). Three participants rate each image, and responses with majority consensus, i.e., min/max scores within 1 point, are included.

\begin{figure}[t!]
    \centering
    \includegraphics[width=\linewidth, height=5cm]{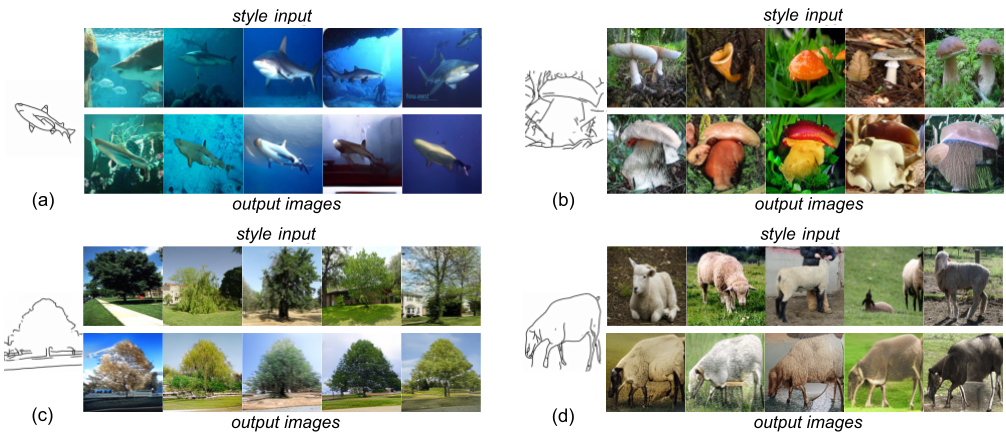}
    \caption{Style controllability.  For each subfigure (a-d), we generate outputs using a single pseudosketch and 5 different style images.}
    \label{fig:style-control}
\end{figure}

We demonstrate style control in \cref{fig:style-control}. We algorithmically select the 5 style images for each sketch by taking the top $N=100$ nearest neighbors within the same class in ALADIN style space to the ground truth image corresponding to the input sketch, clustering the $N$ style vectors into $k=5$ clusters using $k$-means, and finding the image closest to each of the $k$ centroids. We show that the output images capture the variations in the style image.

\textbf{Structure controllability.} To quantitatively measure the structural fidelity of the generated image to the input sketch, we compute the Chamfer distance $d_{structure}(x, e_{\hat{y}}) = \frac{1}{n}\sum_{i \in x} v_i$, where $x$ is the input sketch, $e_{\hat{y}}$ is the edgemap extracted from the generated image $\hat{y}$ using the Canny edge detector \cite{canny1986computational}, and $v_i$ is the distance transform value of $e_{\hat{y}}$. We only sum over the black pixel coordinates $i$ of $x$ to measure the structural coherence of just the target object. In \cref{tab:controllability} we report the mean structure distance and AMT evaluations, where workers are asked to rate how well the generated image matches the input sketch contours on a scale of 1 (worst) to 5 (best) with consensus. 

\begin{figure}[t!]
    \centering
    \includegraphics[width=\linewidth,height=5cm]{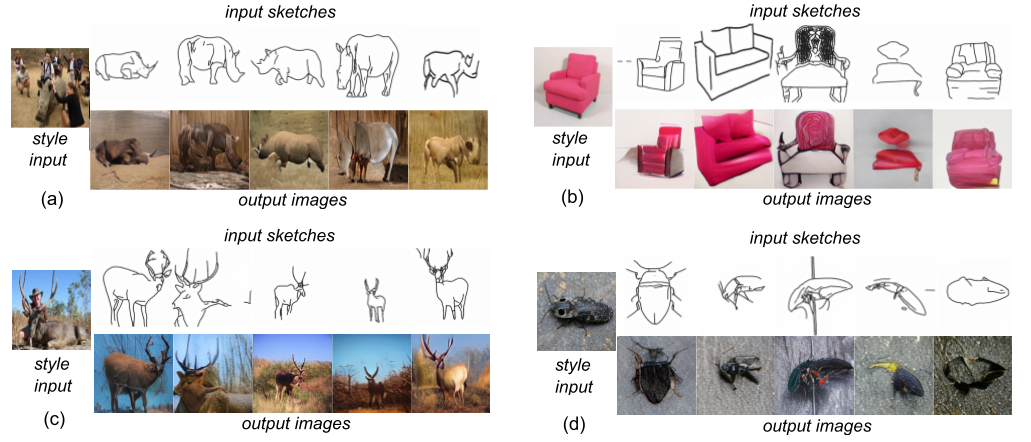}
    \caption{Structure controllability.  For each subfigure (a-d), we generate outputs using 5 pseudosketches for a single style image.}
    \label{fig:structure-control}
\end{figure}

We qualitatively demonstrate the ability to control the structure of the output by sampling the top 10\% of pseudosketches for each class (as determined by the AMT score during data collection), and randomly sample one style image from within the same class. We visualize the results in \cref{fig:structure-control} to show that the synthesized image is guided by the contours of the input sketch.

\subsection{Generalization to hand-drawn sketches} \label{sec:exp-sketchydb}

\begin{figure}[t!]
    \centering
    \includegraphics[width=\linewidth,height=5cm]{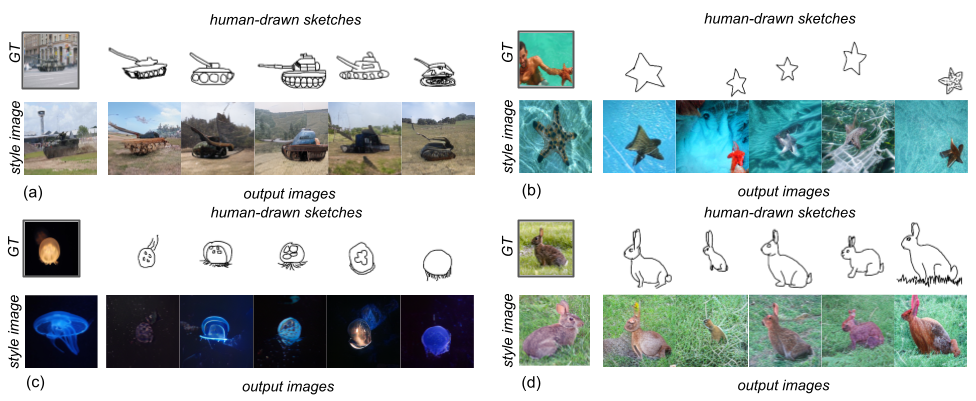}
    \caption{Generalization. For each subfigure (a-d), we generate outputs using 5 sketches from Sketchy DB \cite{sangkloy2016} corresponding to a ground truth image (framed in grey) with a given style image to demonstrate generalization to hand-drawn sketches.}
    \label{fig:sketchydb}
\end{figure}

Pseudosketches are similarly minimalistic to human-drawn sketches, but their contours have direct pixel correspondences to their ground truth images, whereas human-drawn sketches may come from users of varying skill levels and are more abstract. Although CoGS is only trained on pseudosketches, we show that it is able to generalize to some of the higher quality human-drawn sketches from Sketchy DB (see \cref{fig:sketchydb}). Because there is a strong correlation in the number of examples used to compute the FID, we first randomly sample a subset of Sketchy DB similar in size to the Pseudosketches validation set. Our method achieves an FID of 81.820 on Sketchy DB, compared to 50.630 on Pseudosketches, highlighting a gap in the quality of synthesis results from using hand-drawn sketches compared to pseudosketches due to the domain gap and abstraction.

\subsection{Transformer ablation study} \label{sec:exp-ablation}

We investigate the impact of various components of the CoGS transformer using FID, LPIPS, style distance, and structure distance. We show in \cref{tab:ablation} that both the class label $c$ and style loss $\mathcal{L}_{style}$ are important in generating a diverse set of realistic outputs that are consistent with the stylistic and structural conditions.

\begin{figure}[t!]
    \centering
    \includegraphics[width=\linewidth,height=6cm,trim=0cm 2cm 0cm 0cm,clip]{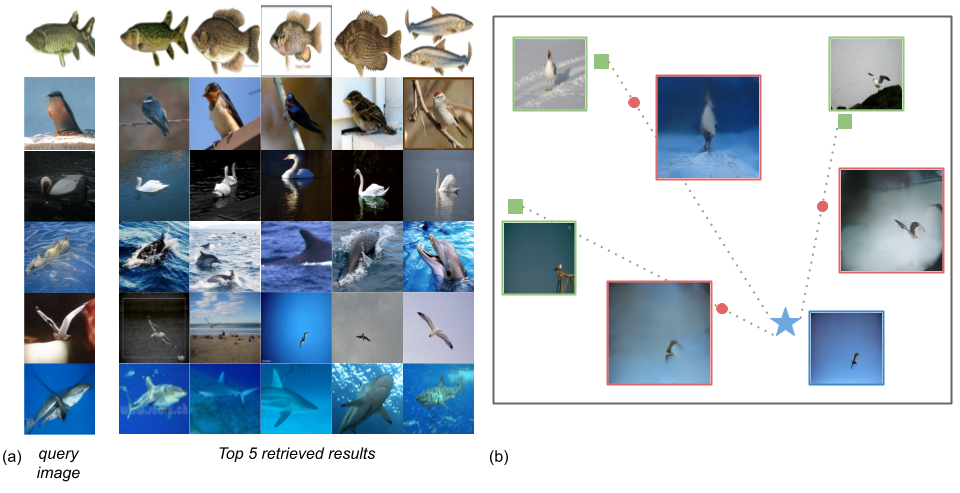}
    \caption{(a) Top 5 images retrieved from the validation set using generated images as queries. (b) Depiction of the latent space exploration for synthesizing new results. Blue $\star$: query image, green $\blacksquare$: retrieval results, red $\bullet$: interpolation results.}
    \label{fig:retrievalresults}
\end{figure}

\begin{table}[h!]
\centering
\begin{adjustbox}{width=0.9\textwidth}
\begin{tabular}{cccccc}
\toprule
\textbf{Inputs}  &  \textbf{Losses}  &  \textbf{FID}$\downarrow$  &  \textbf{LPIPS}$\uparrow$  &  \textbf{Style}$\downarrow$  &  \textbf{Structure}$\downarrow$ \\ \cmidrule{1-6}
sketch, style  &  codebook  &  58.202  &  0.519  &  1.117 &  3.063  \\
sketch, style  &  codebook, style  &  58.327  &  0.514  &  1.129  &  2.823  \\
sketch, style, label  &  codebook  &  52.992  &  0.518  &  1.120  &  3.043  \\
sketch, style, label  &  codebook, style  &  50.630  &  0.521  &  1.100  &  2.387  \\ \bottomrule
\end{tabular}
\end{adjustbox}
\caption{Ablation study on the inputs and losses for our proposed method. }

\label{tab:ablation}
\end{table}

\subsection{Fine-grained control via image retrieval and interpolation} \label{sec:exp-refinement}

We first evaluate the latent space at offering users similar images, given the transformer output and its class label. We use a VAE trained on the query image class and retrieve the closest validation images from the same class. We show in \cref{fig:retrievalresults}(a) how, independent of the style and quality of the query image, the retrieved images have the most similar structure. This behavior is validated by AMT evaluations, where 3 participants were asked to rate each retrieved image of a synthesized query image as relevant or not. The majority vote was used to calculate precision@$k$ of $0.533$, $0.535$, $0.530$, $0.520$, $0.425$ at $k=\{1, 5, 10, 15, 20\}$, respectively, showing how workers agreed on the relevance of up to around the top 15 images, which demonstrates local coherency. Next, we consider the top 3 retrieved images and synthesize new ones by interpolating between each of retrieved images and the query image. We sample 50 images for each interpolation pair, and filter them using an FID threshold of 120 to ensure the quality of the results. We depict the interpolation scheme in \cref{fig:retrievalresults}(b), showing a few of the many variations the user can create by mixing attributes from two images.

\section{Conclusion}

We introduce CoGS, a method for image synthesis across a diverse set of categories that provides control over the style and structure of the output image. In order to learn effective controllable synthesis, we collect a large-scale dataset of ``pseudosketch''-image correspondences using an automated pipeline. Our approach produces images with higher fidelity to the given style and structure constraints, while also producing diverse and more realistic images within those conditions. We also learn a unified embedding for search and synthesis, which enables further refinement of the generated images via retrieval or interpolation. 

\textbf{Limitations.} To broaden the application of our method, we would like to use inputs of hand-drawn sketches from a wide range of skill levels. As shown in \cref{sec:exp-sketchydb}, we are able to generalize to examples of hand-drawn sketches with better artistry, but there is still a gap in the FID computed on sketches across skill levels. While this is understandable due to domain gap from the pseudosketches training data, there is room for improvement in resolving abstraction of structure present in hand-drawn sketches.


\bibliographystyle{splncs04}
\bibliography{main}


\newpage

\appendix

\section{Pseudosketches Dataset}

\begin{figure}[]
  \centering
  \includegraphics[width=\textwidth]{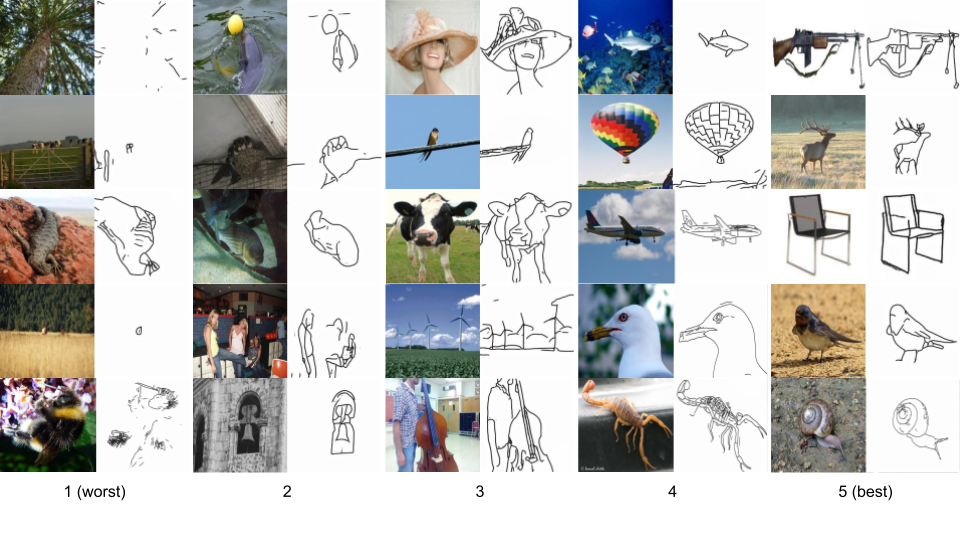}
  \caption{Examples of image-pseudosketch pairs at each AMT score.}
  \label{fig:examples-per-score}
\end{figure}

To build the Pseudosketches dataset, we start with the 12.5K images from Sketchy DB \cite{sketchydb}, and sample additional images for each category (or similar categories based on WordNet distances to the original category) from ImageNet \cite{deng2009imagenet}. After running the automated pseudosketch-extraction pipeline on all gathered images, we display the pseudosketch-image pair and its class label, and ask Amazon Mechanical Turk (AMT) workers to rate how well the pseudosketch represents its corresponding image on a scale of 1 (worst) to 5 (best). We visualize paired examples for each AMT score in \cref{fig:examples-per-score}.

\begin{figure}[]
  \centering
  \includegraphics[height=\textheight,trim=0cm 1cm 0cm 2cm,clip]{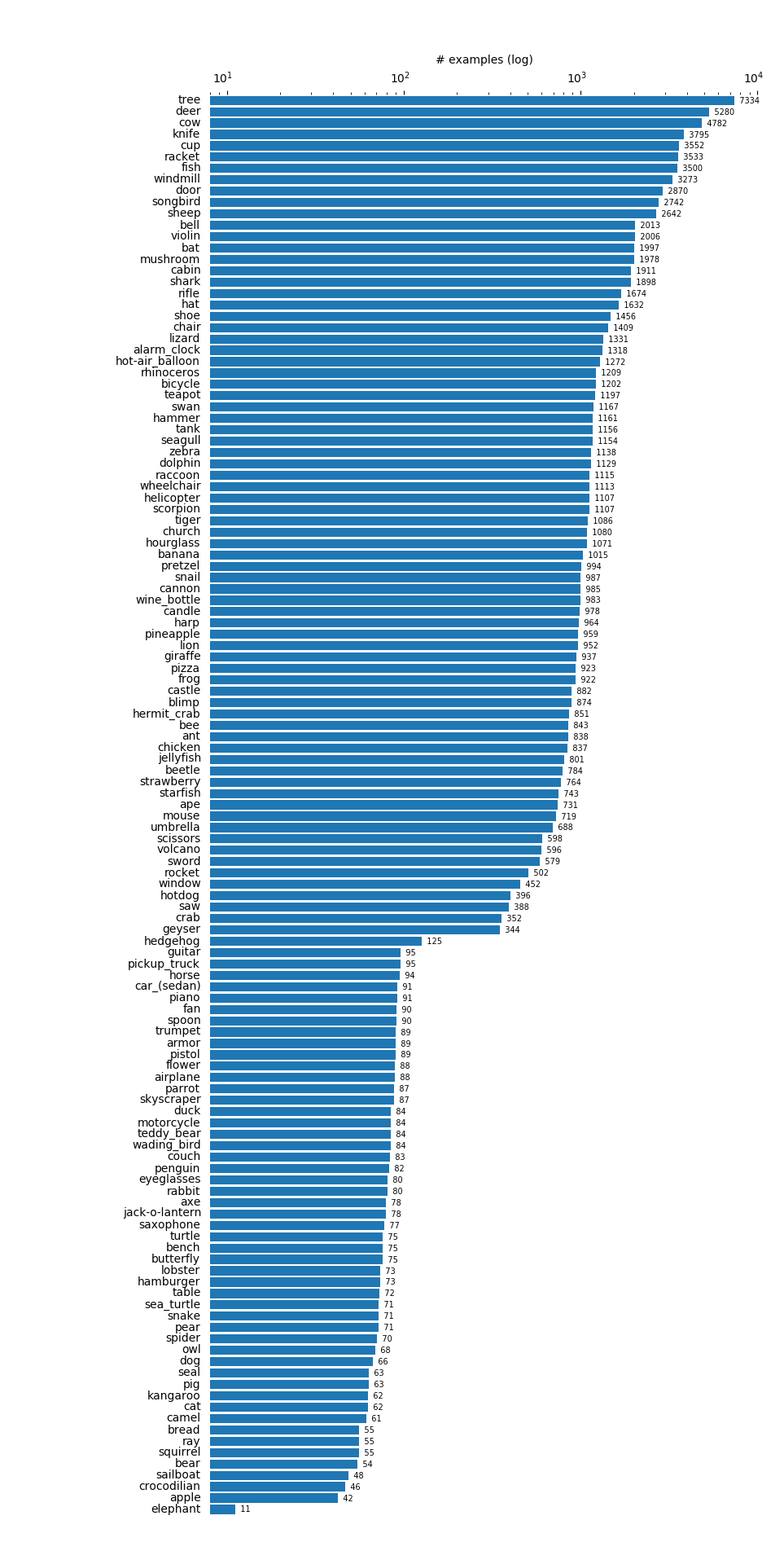}
  \caption{Number of images per category in the Pseudosketches dataset.}
  \label{fig:dataset-distribution}
\end{figure}

After removing pseudosketches with a score of 1 or 2, we are left with 113,700 pseudosketches across the original 125 categories of Sketchy DB (see \cref{fig:dataset-distribution} for the distribution of pseudosketches per category). We note that the dataset is imbalanced due to some categories having more images than others and/or some categories producing many more poorly-scored pseudosketches than others.

\section{Image synthesis using transformers}
\subsection{Comparison to CNN-based networks}

\begin{figure}[]
  \centering
  \includegraphics[width=0.7\textwidth]{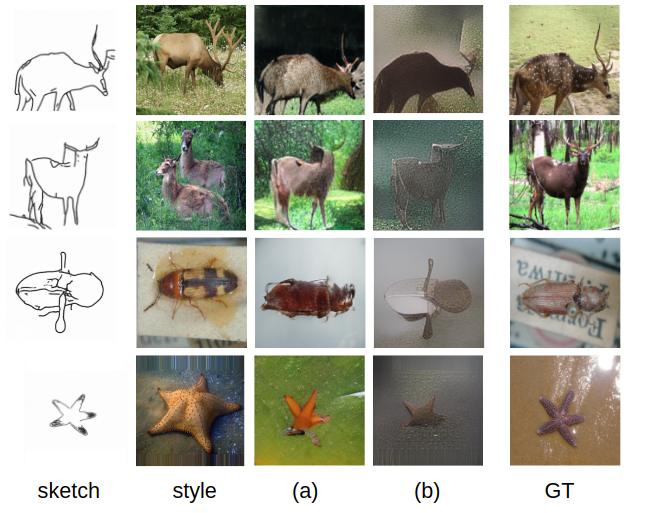}
  \caption{For a given labeled sketch and style image, (a) is the corresponding synthesized image using transformers, and (b) represents the synthesized image using a StyleGAN2-based network. The rightmost column is the ``ground truth'' image.}
  \label{fig:cnn-synthesis}
\end{figure}

We explore the use of various CNN-based networks instead of transformers for synthesis. The most successful alternative used a decoder adapted from StyleGAN2 \cite{karras2019stylegan2} to learn the composite codebook representation given a set of conditional inputs, but we found the network to be susceptible to overfitting. Even when additional information was given to the CNN, such as pseudosketch of the style image, transformers proved to better interpret and learn the distribution of codebooks, leading to better control and more high frequency details (see \cref{fig:cnn-synthesis}).

\subsection{Qualitative ablation study}

\begin{figure}[]
  \centering
  \includegraphics[width=\textwidth]{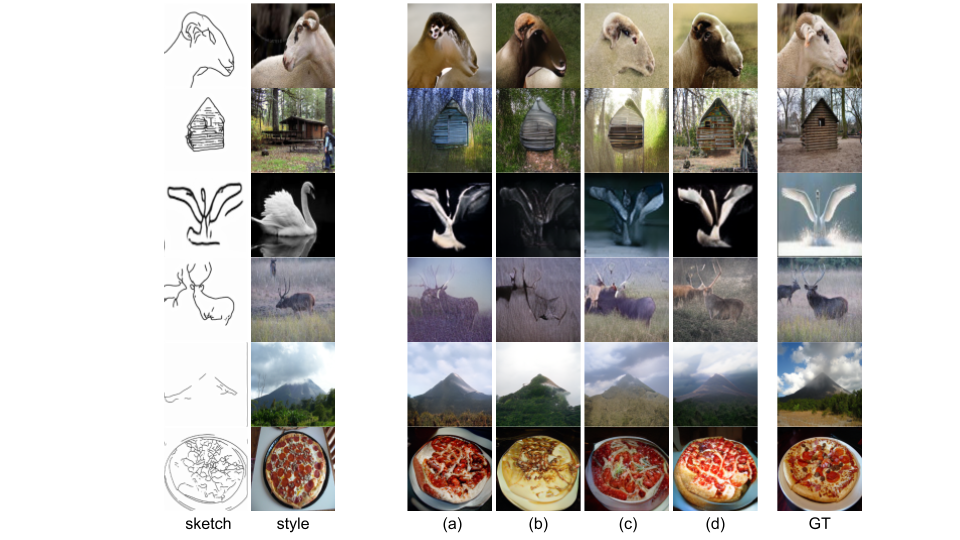}
  \caption{Visualization of synthesized images from the different ablated methods: (a) \textit{inputs:} sketch, style image, \textit{losses:} codebook, (b) \textit{inputs:} sketch, style image, \textit{losses:} codebook, style, (c) \textit{inputs:} sketch, style image, class label, \textit{losses:} codebook, (d) \textit{inputs:} sketch, style image, class label, \textit{losses:} codebook, style.}
  \label{fig:ablationfigure}
\end{figure}

In \cref{fig:ablationfigure} we present qualitative results of the ablation study performed in the main paper to demonstrate the importance of the class label and auxiliary style loss for enabling control over the output image. Specifically, adding the class label makes the generated object more faithful to its semantic structure and texture, while the additional loss allows for a better style control over the output.

\subsection{Data partitioning}

\begin{figure}[]
  \centering
  \includegraphics[width=\textwidth]{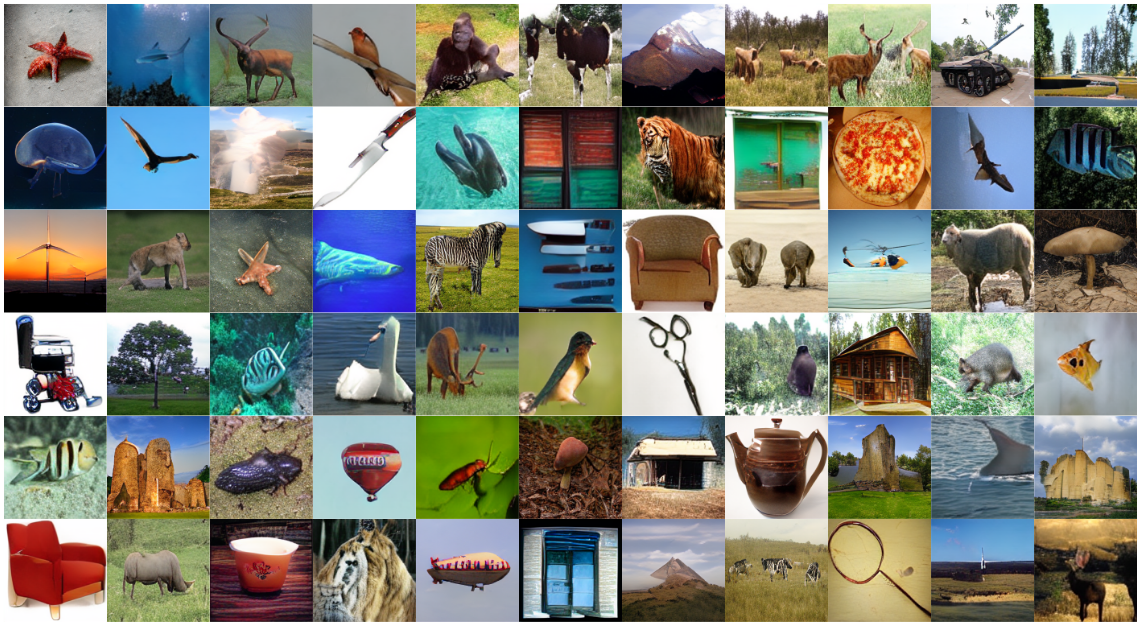}
  \caption{Examples of synthesizes images for all classes in the ``simple'' partition.}
  \label{fig:simplepartition}
\end{figure}

\begin{figure}[]
  \centering
  \includegraphics[width=\textwidth]{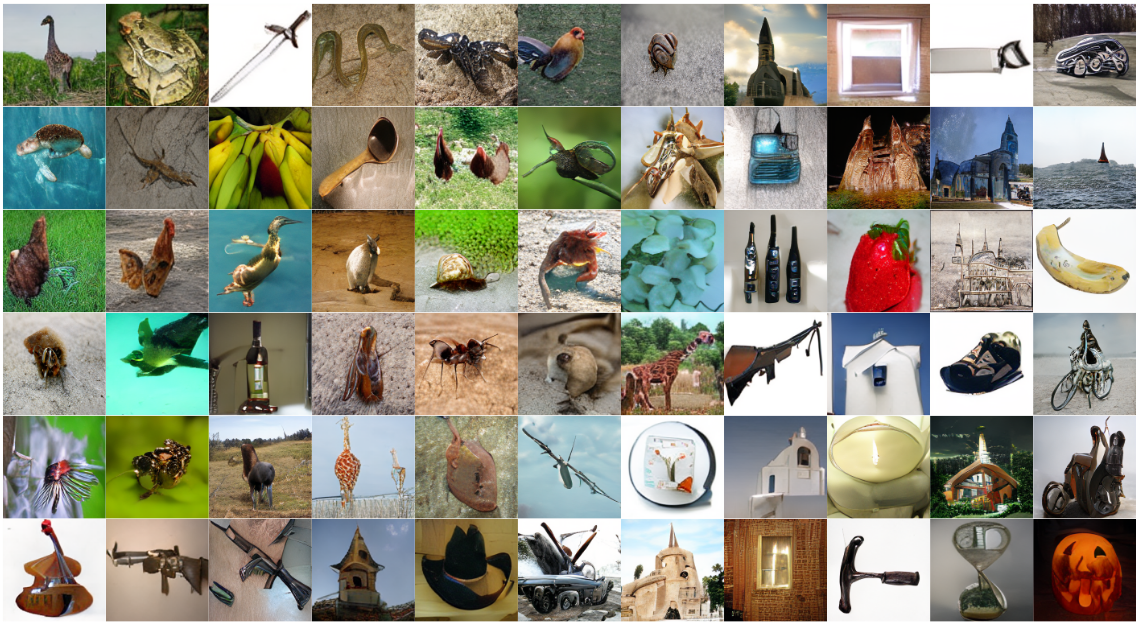}
  \caption{Examples of synthesized images for all classes in the ``medium'' partition.}
  \label{fig:medpartition}
\end{figure}

\begin{figure}[]
  \centering
  \includegraphics[width=\textwidth]{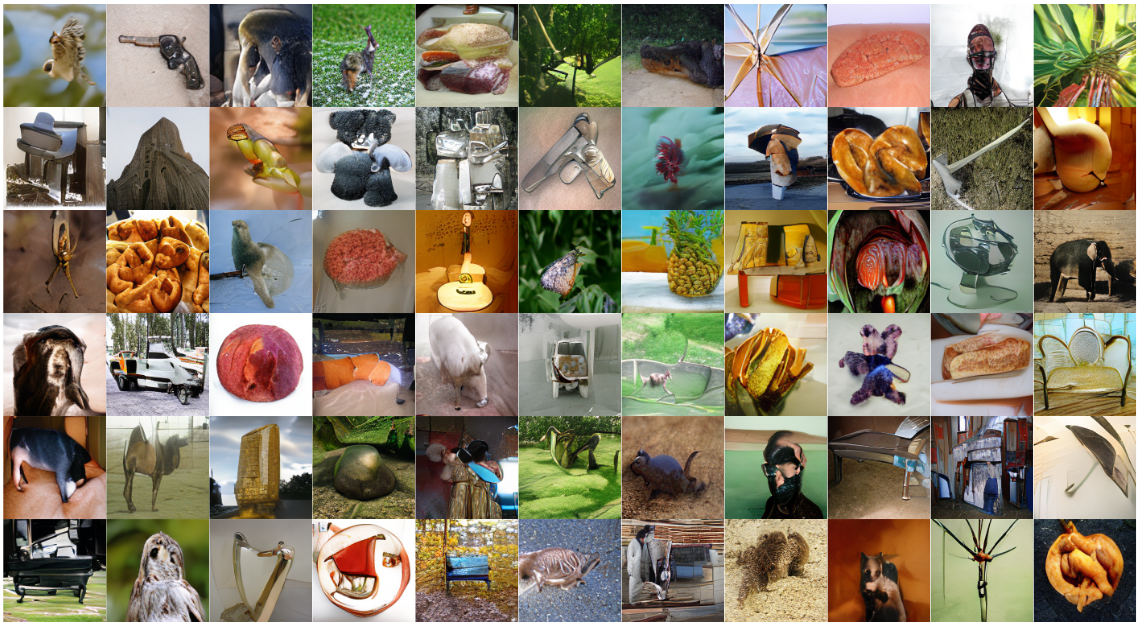}
  \caption{Examples of synthesized images for all classes in the ``complex'' partition.}
  \label{fig:complexpartition}
\end{figure}

We partition the categories of the Pseudosketches dataset by computing class-wise FID \cite{heusel2017gans} scores with the validation images in each of the 125 categories (see \cref{tab:partition-classes}). \cref{fig:simplepartition}, \cref{fig:medpartition}, and \cref{fig:complexpartition} visualize examples from each of the ``simple'', ``medium'', and ``complex'' partitions, respectively. We observe common characteristics of the categories belonging to the more difficult partitions, such as less uniform textures and multiple non-related objects or humans present. However, we are still able to demonstrate the ability of CoGS to synthesize a diverse set of categories, using semantic understanding to generate appropriate textures and even applying realistic lighting, shadows, and reflections.

\begin{table}[]
  \centering
  \begin{tabular}{cc}
    \toprule
    \textbf{Partition}  &  \textbf{Categories}  \\
    \cmidrule{1-2}
    \multirow{5}{*}  &  deer, tree, cow, zebra, songbird, windmill, door, shark, rhinoceros, \\ 
    &  cabin, cup, knife, sheep, dolphin, chair, seagull, swan, castle, pizza \\
    Simple  &  fish, volcano, mushroom, beetle, lion, hot-air balloon, bat, ape, tiger, \\
    &  helicopter, teapot, wheelchair, geyser, scissors, starfish, tank, \\
    &  jellyfish, rocket, raccoon, blimp, racket, wading bird \\
    \cmidrule{1-2}
    \multirow{5}{*}  &  snail, church, giraffe, sword, jack-o-lantern, lizard, sailboat, car (sedan), \\
    &  bicycle, rifle, ant, saw, bee, window, frog, alarm clock, shoe, bell, \\
    Medium  &  scorpion, hermit crab, ray, hat, wine bottle, hourglass, spoon, motorcycle, \\
    &  penguin, sea turtle, candle, hammer, chicken, snake, kangaroo, strawberry, \\
    &  duck, violin, airplane, banana, cannon, crab, mouse, horse \\
    \cmidrule{1-2}
    \multirow{5}{*}  &  hot dog, pineapple, owl, butterfly, pretzel, rabbit, hedgehog, pear, \\
    &  pistol, umbrella, hamburger, bear, bench, camel, parrot, fan, pig, \\
    Complex  &  pickup truck, table, apple, seal, elephant, armor, spider, flower, squirrel, \\
    &  piano, bread, turtle, eyeglasses, guitar, crocodilian, axe, skyscraper, \\
    &  couch, cat, teddy bear, trumpet, dog, saxophone, harp, lobster \\
    \bottomrule
  \end{tabular}
  \caption{Categories belonging to each of the 3 partitions.}
  \label{tab:partition-classes}
\end{table}

\subsection{Style and structure controllability}

\begin{figure}[]
  \centering
  \includegraphics[width=0.65\textwidth]{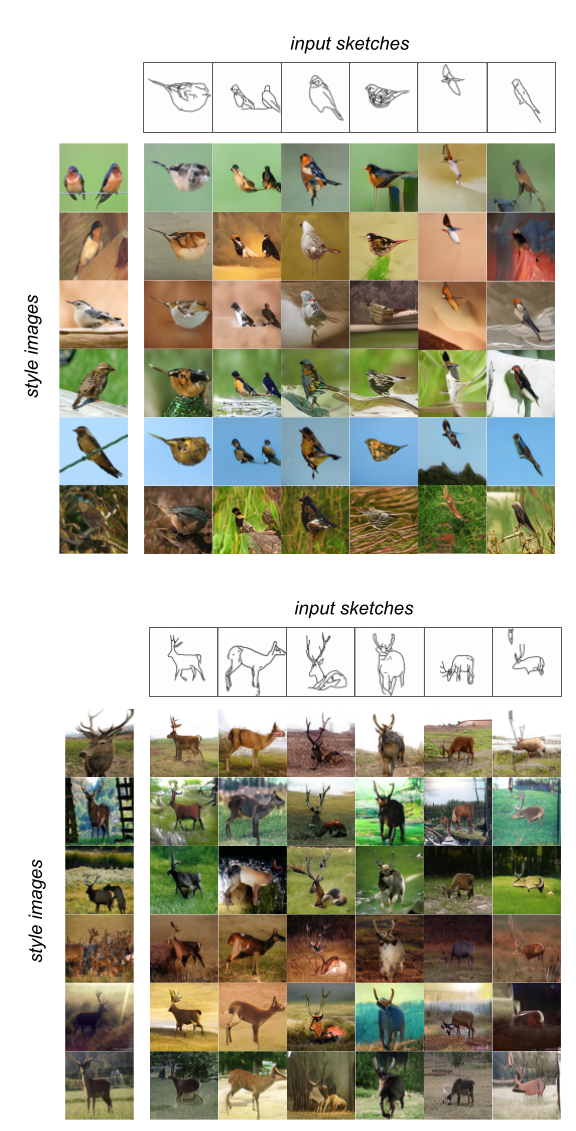}
  \caption{Images synthesized by CoGS using the respective row and column input combination for the \textit{songbird}  and \textit{deer} classes.}
  \label{fig:matrixbirddeer}
\end{figure}

\begin{figure}[]
  \centering
  \includegraphics[width=0.65\textwidth]{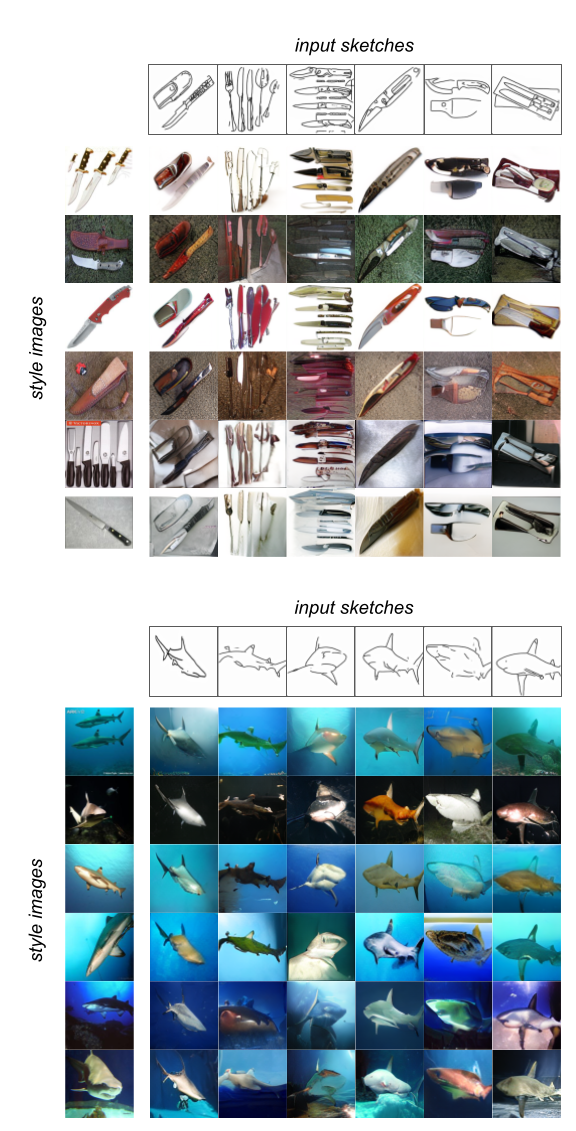}
  \caption{Images synthesized by CoGS using the respective row and column input combination for the \textit{knife}  and \textit{shark} classes.}
  \label{fig:matrixknifeshark}
\end{figure}

We propose CoGS as an image synthesis method that provides decoupled control over the structure and style of the generated image through sketch and style image inputs, respectively. In \cref{fig:matrixbirddeer} and \ref{fig:matrixknifeshark} we vary the two axes of control to show that our method captures both the structure of the input sketch and the style of the input style image.

\subsection{Generalization to hand-drawn sketches}

\begin{figure}[]
  \centering
  \begin{subfigure}
    \centering
    \includegraphics[width=\textwidth]{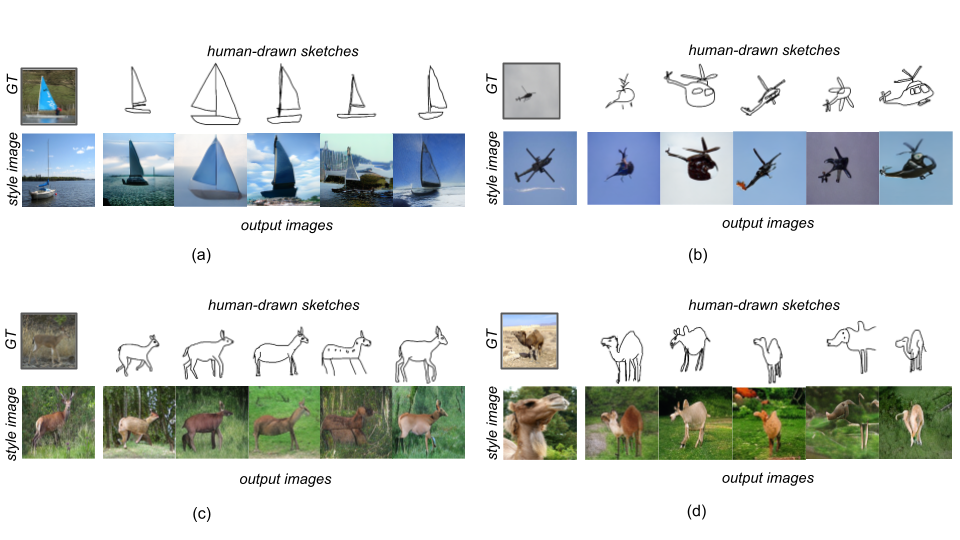}
  \end{subfigure}
  \begin{subfigure}
    \centering
    \includegraphics[width=\textwidth]{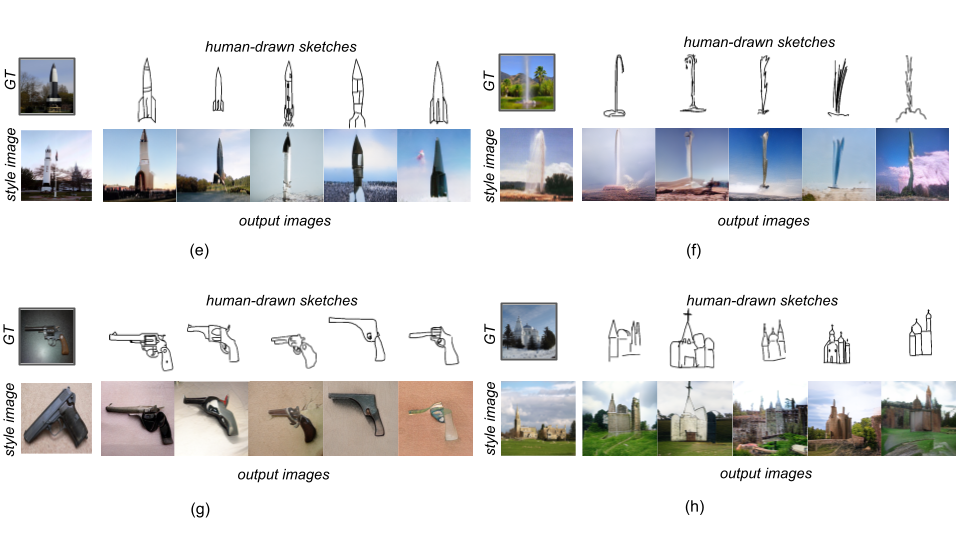}
  \end{subfigure}
  \caption{For each subfigure (a-h), we generate the output using 5 hand-drawn sketches from Sketchy DB \cite{sketchydb} corresponding to a ground truth image (framed in grey) with a given style image.}
  \label{fig:sketchy}
\end{figure}

We demonstrate the ability of CoGS, trained only on pseudosketches, to generalize to the higher quality human-drawn sketches from Sketchy DB \cite{sketchydb} (see \cref{fig:sketchy}), which are more abstract and less faithful to the contours of their corresponding image prompt. Because Sketchy DB was collected from users of varying skill levels, we are able to see the impact of the artistry level on the outputs, highlighting that there still exists a gap between the synthesis quality of the two types of sketch inputs. 

\section{Image refinement using VAEs}
\subsection{Latent space visualization}

\begin{figure}[]
  \centering
  \includegraphics[width=\textwidth]{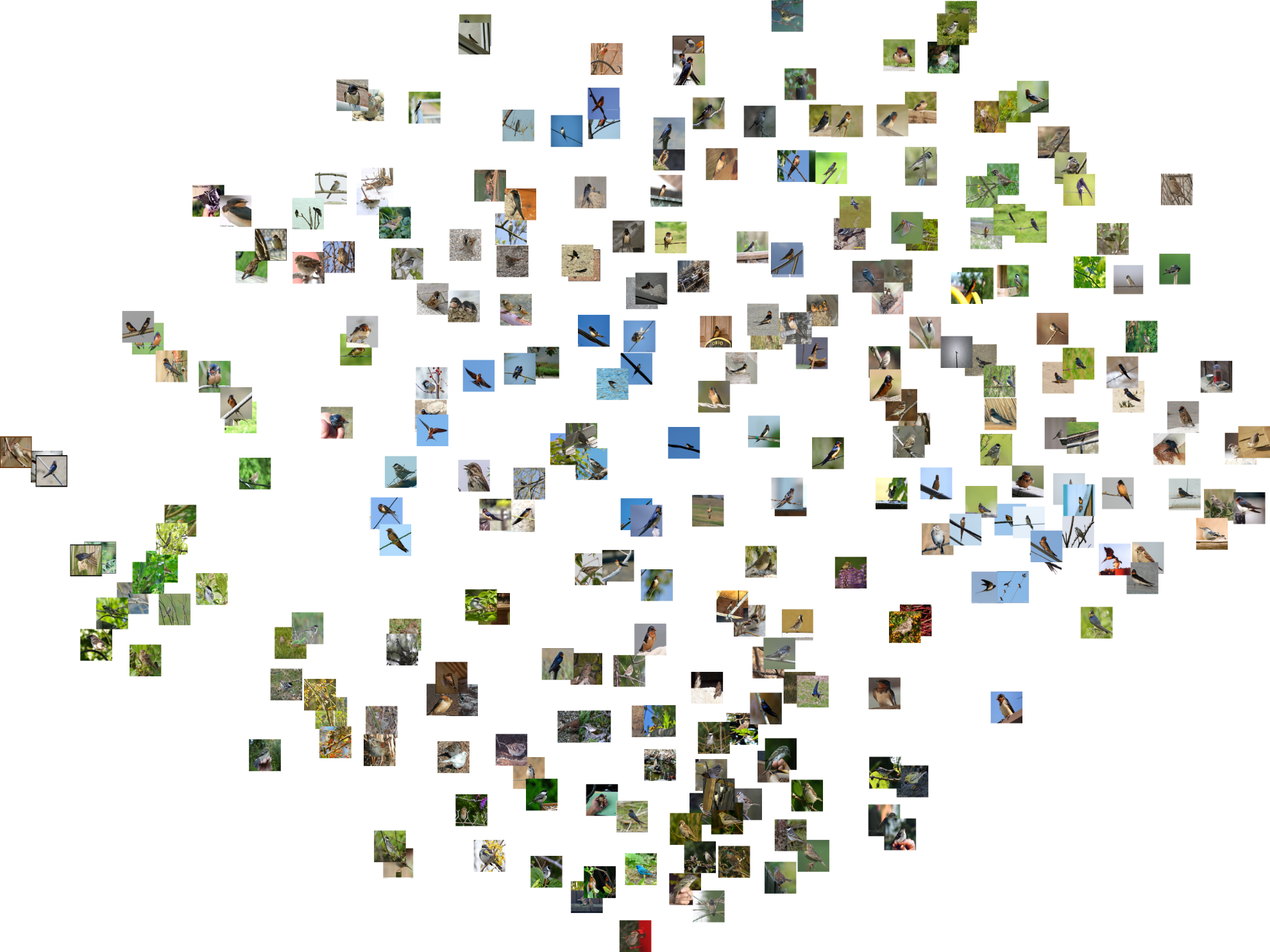}
  \caption{t-SNE visualization of the \textit{songbird} latent space.}
  \label{fig:tsnebird}
\end{figure}

\begin{figure}[]
  \centering
  \includegraphics[width=\textwidth]{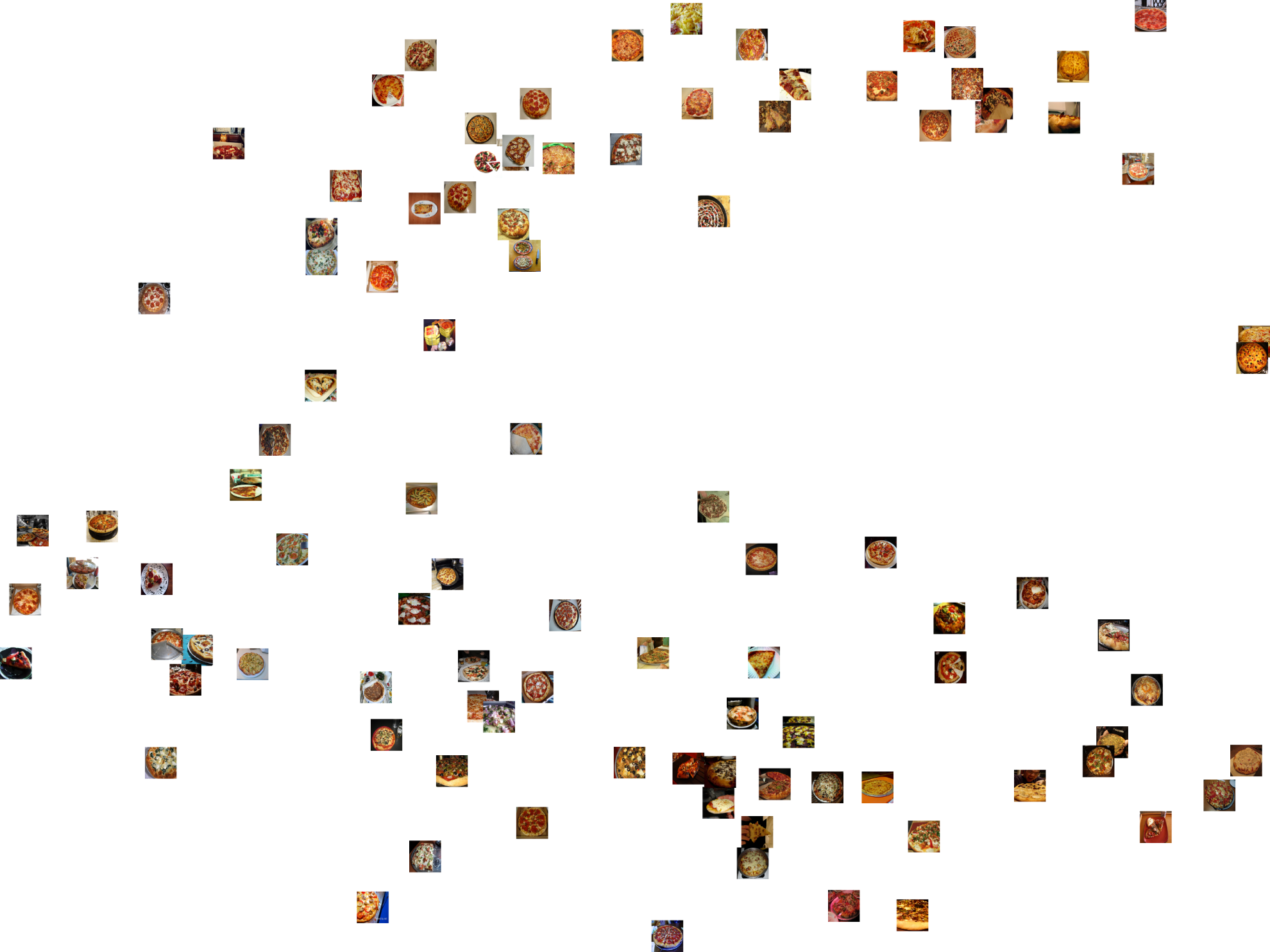}
  \caption{t-SNE visualization of the \textit{pizza} latent space.}
  \label{fig:tsnepizza}
\end{figure}

\begin{figure}[]
  \centering
  \includegraphics[width=\textwidth]{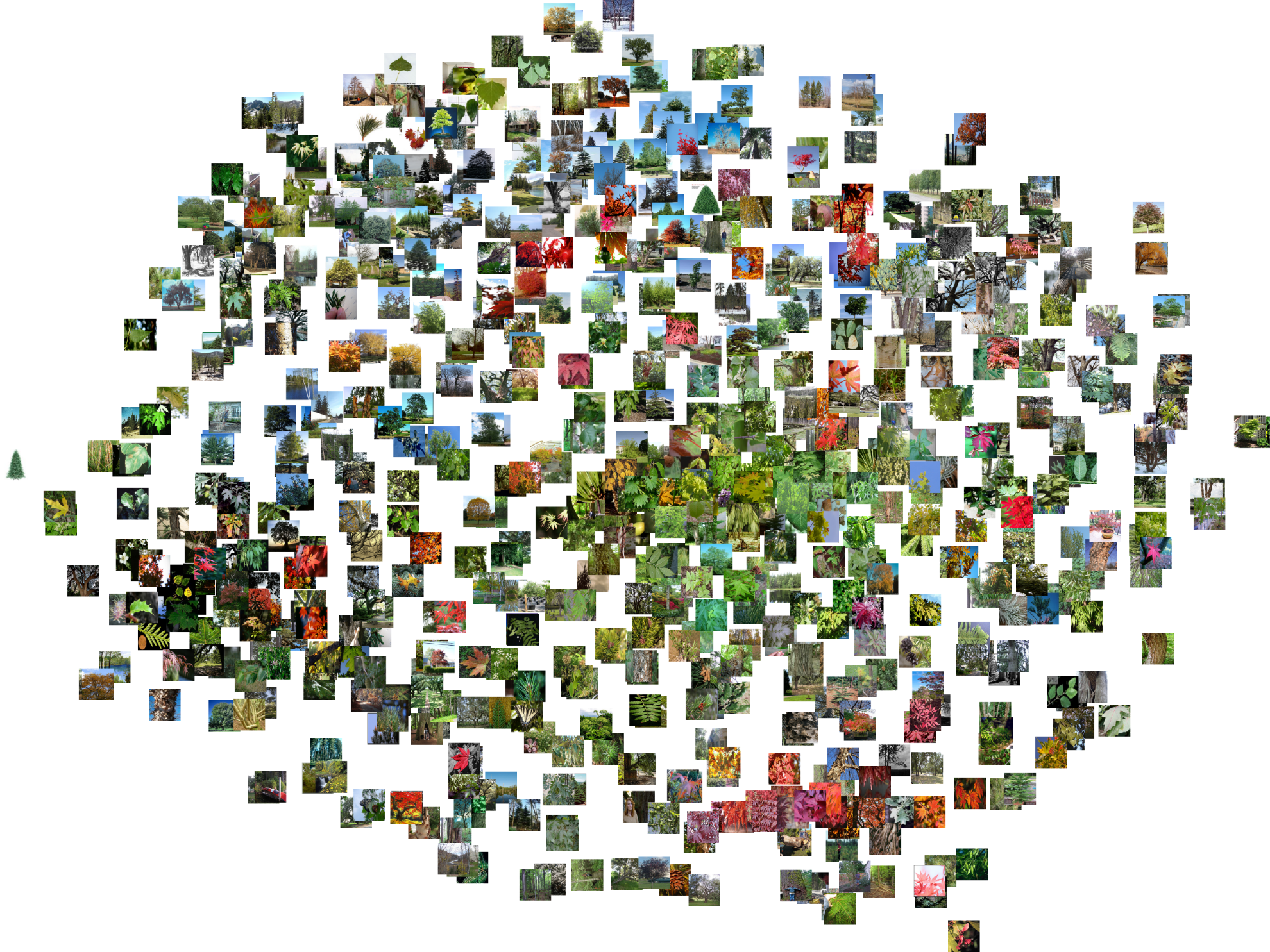}
  \caption{t-SNE visualization of the \textit{tree} latent space.}
  \label{fig:tsnetree}
\end{figure}

CoGS offers an optional step to refine the generated output of the transformer through the use of variational autoencoders (VAE) \cite{VAE}. We train a VAE for each class and visualize the structure of a few latent spaces using t-Distributed Stochastic Neighbor Embedding (t-SNE) \cite{van2008visualizing} on the Pseudosketches validation set in \cref{fig:tsnebird}, \cref{fig:tsnepizza}, and \cref{fig:tsnetree}. We observe that the contrastive training paradigm yields latent spaces in which images with similar structures are closer together, and images with dissimilar structures are further apart.

\subsection{Photorealistic image retrieval}

\begin{figure}[]
  \centering
  \includegraphics[width=\textwidth]{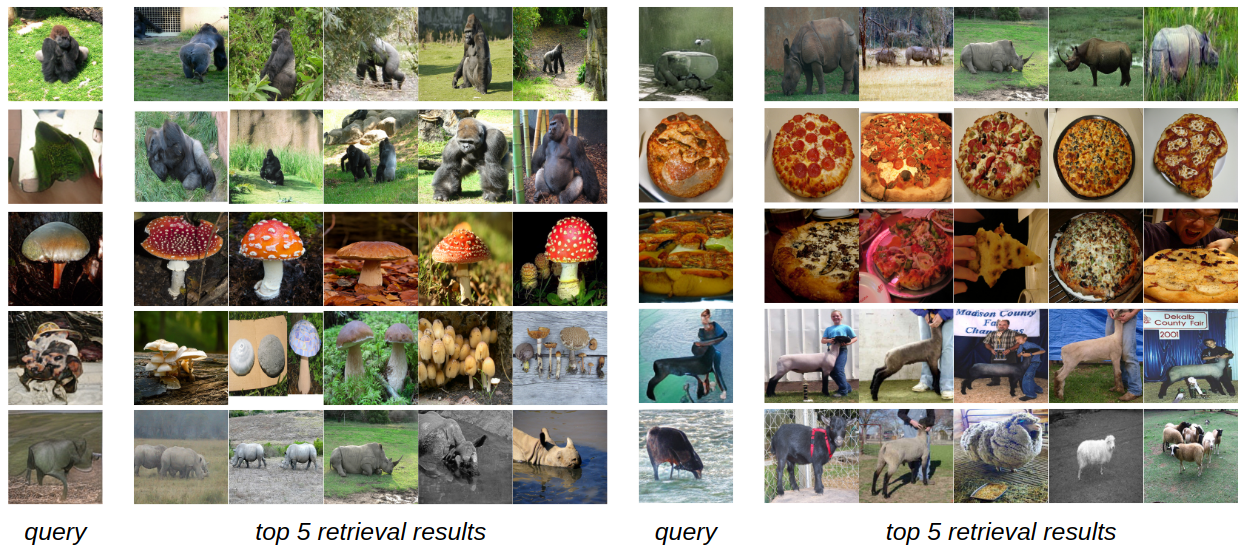}
  \caption{Top 5 retrieval results for query images from the ``simple'' partition.}
  \label{fig:retrievalsimple}
\end{figure}

\begin{figure}[]
  \centering
  \includegraphics[width=\textwidth]{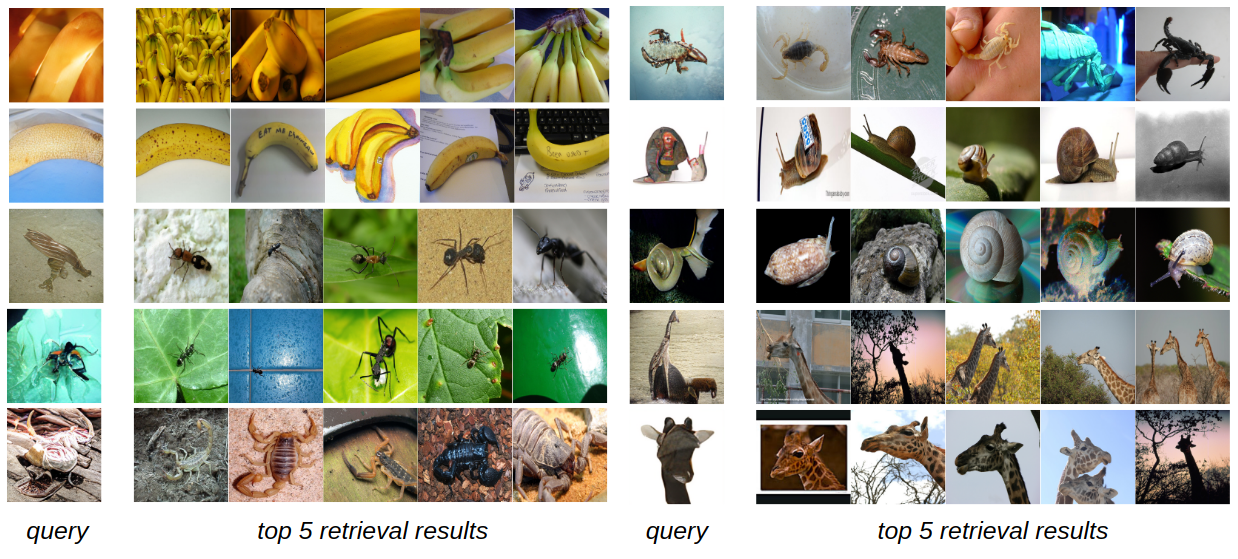}
  \caption{Top 5 retrieval results for query images from the ``medium'' partition.}
  \label{fig:retrievalmed}
\end{figure}

\begin{figure}[]
  \centering
  \includegraphics[width=\textwidth]{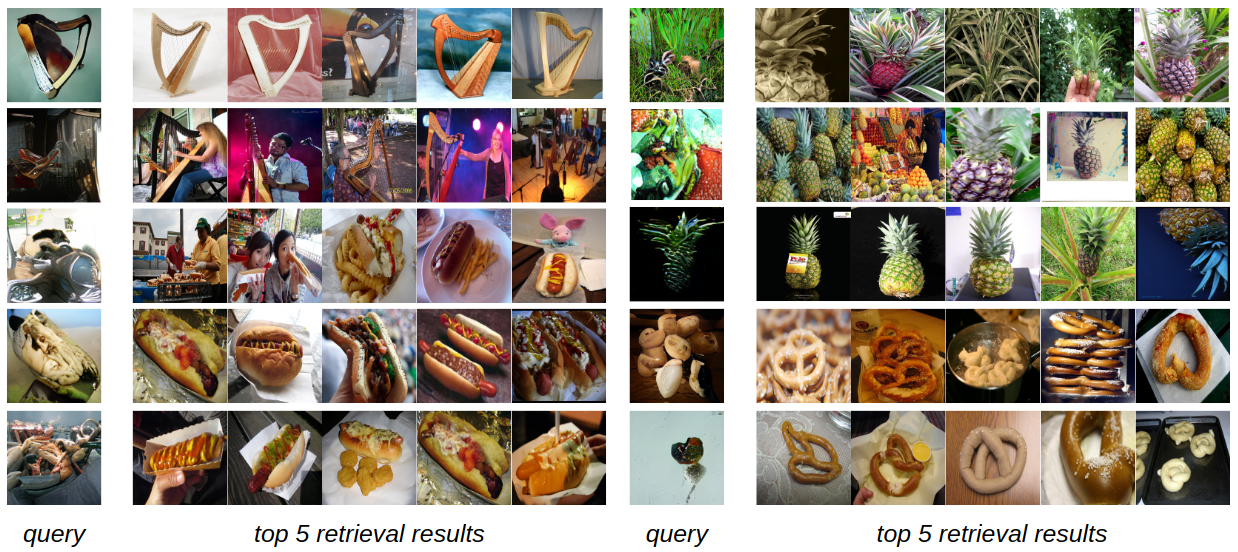}
  \caption{Top 5 retrieval results for query images from the ``complex'' partition.}
  \label{fig:retrievalhard}
\end{figure}

We study the difference in retrieval performance across the three partitions of the Pseudosketches dataset by sampling images generated by CoGS and retrieving the top 20 photorealistic images within the same class from the Pseudosketches dataset. Through AMT evaluations we evaluate the precision@$k$ for the top 20 retrieved images (see \cref{tab:retrieval}), and show coherence of the latent space and relevancy of the retrieved results. We visualize retrieval results for queries belonging to each of the partitions in \cref{fig:retrievalsimple}, \cref{fig:retrievalmed}, and \cref{fig:retrievalhard}.

\begin{table}[t]
  \centering
  \begin{tabular}{cccccc}
    \toprule
    \textbf{Partition}  &  \textbf{$k=1$}  &  \textbf{$k=5$}  &  \textbf{$k=10$}  &  \textbf{$k=15$} & \textbf{$k=20$}\\
    \cmidrule{1-6}
    Simple  &  0.856  &  0.735  &  0.744  &  0.769 & 0.770 \\
    Medium  &  0.931  & 0.910  &  0.898  &  0.901 & 0.892 \\
    Complex  &  0.935 &  0.937  &  0.702  &  0.941 & 0.848 \\
    \bottomrule
  \end{tabular}
  \caption{Precision@$k$ with $k=\{1,5,10,15,20\}$ for the retrieved results across various classes within each partition.}
  \label{tab:retrieval}
\end{table}

\subsection{Latent space interpolation}

\begin{figure}[]
  \centering
  \includegraphics[width=\textwidth]{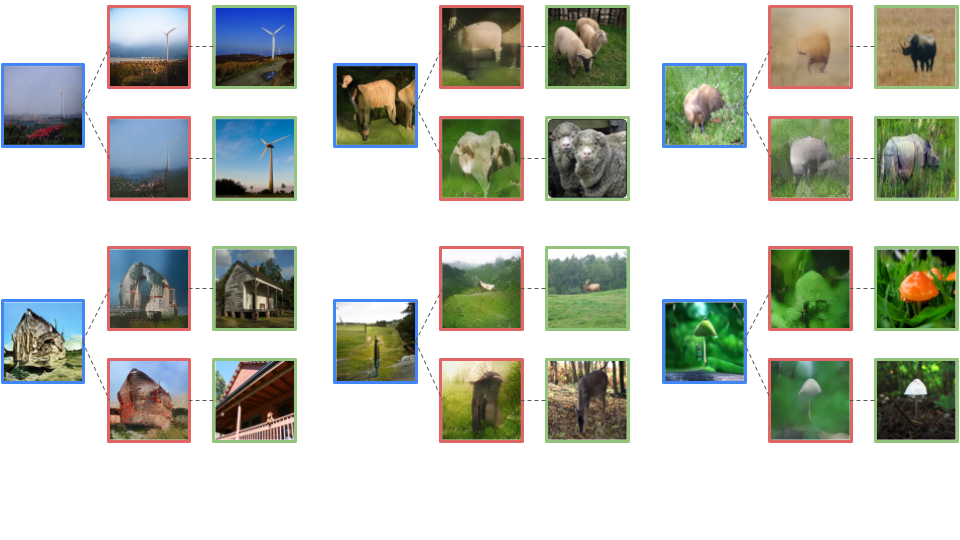}
  \caption{For a given query image (blue box) we retrieve two of its nearest photorealistic neighbors (green) and synthesize images (red) by interpolating between the query and each retrieval result.}
  \label{fig:interpolation}
\end{figure}

In \cref{fig:interpolation} we visualize images synthesized by interpolating between query images and their top retrieval results from the Pseudosketches dataset.

\subsection{Sketch-based image retrieval}

\begin{figure}[t!]
  \centering
  \includegraphics[width=0.9\linewidth,height=6cm]{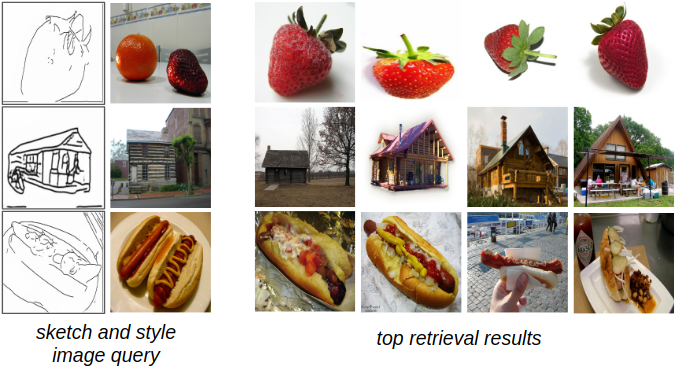}
  \caption{Top 4 images retrieved by using CoGS as a SBIR method with (sketch, style) pairs as inputs.}
  \label{fig:sbir}
\end{figure}

CoGS accepts a (sketch, style, label) input for synthesis via a codebook representation, which may be used as a query for retrieval. The output images may be interpolated or used directly to refine the synthesized image. While sketch-based image retrieval (SBIR) is not the intent of this work, it is possible if a style image is provided (\cref{fig:sbir}).

\section{Efficiency}

CoGS only requires a feed-forward inference pass for initial synthesis, and again for subsequent retrieval/interpolations for refinement (3-5s on a single Titan X).

\section{AMT evaluations}

\begin{figure}[]
  \centering
  \includegraphics[width=\textwidth]{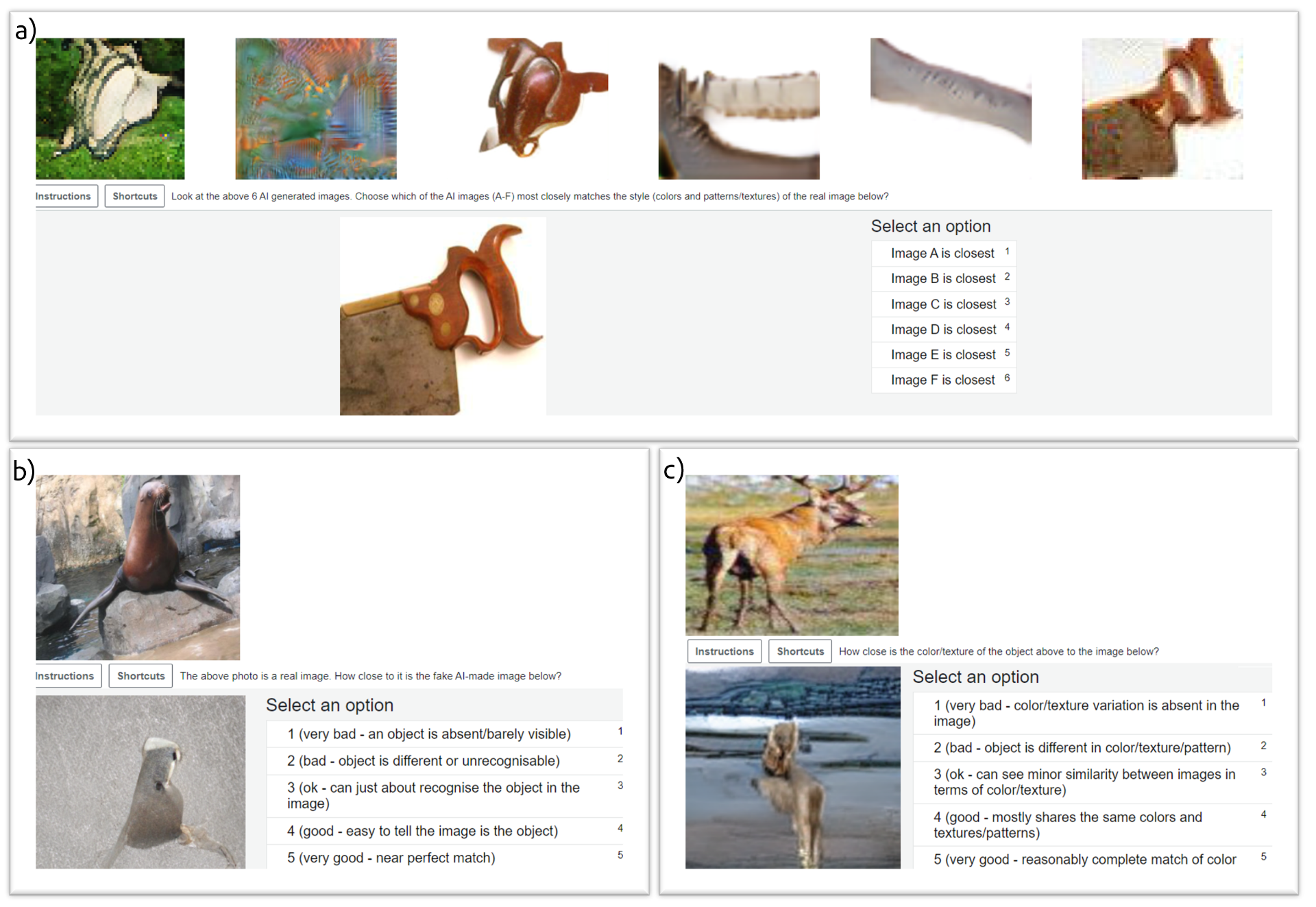}
  \caption{Examples of AMT evaluation task prompts. (a) Prompt for selecting the preferred reconstructed image based on style. (b) Prompt for evaluating the fidelity of each generated image. (c) Prompt for evaluating style controllability on the images generated by CoGS.}
  \label{fig:AMTexample}
\end{figure}

We provide additional details about the three Amazon Mechanical Turk (AMT) tasks (visualized in \cref{fig:AMTexample}) used to crowd-source human evaluations of CoGS and the baseline methods:

\begin{enumerate}
  \item{Baseline comparison (Section 4.3)
    \begin{enumerate}
      \item \textit{Preference based on ground-truth}. ``Look at the above 6 AI generated images. Choose which of the AI images (A-F) most closely matches the REAL image below?'' (select: A-F)
      \item \textit{Preference based on style}. ``Look at the above 6 AI generated images. Choose which of the AI images (A-F) most closely matches the style (colors and patterns/textures) of the real image below?'' (select A-F)
      \item \textit{Preference based on structure}. ``Look at the above 6 AI generated images. Choose which of the AI images (A-F) most closely matches the shape of the sketch below?'' (select A-F)
      \item \textit{Realism}. ``How realistic does the below AI generated image look?'' (select 1 (very bad) - 5 (very good))
      \item \textit{Fidelity}. ``The above photo is a real image. How close to it is the fake AI-made image below?'' (select 1 (very bad) - 5 (very good))
    \end{enumerate}
  }
  \item{Controllability experiments (Section 4.4)
    \begin{enumerate}
      \item \textit{Structure}. ``How close is the shape of the object in the image to the sketched shape?'' (select 1 (very bad) - 5 (very good))
      \item \textit{Style}. ``How close is the color/texture of the object above to the image below?'' (select 1 (very bad) - 5 (very good))
    \end{enumerate}
  }
  \item{Retrieval experiments (Section 4.7)
    \begin{enumerate}
      \item \textit{Retrieval relevance}. ``Examine this image pair. Do both the structure (the shape) and the appearance (colour and texture of the image) of the two images match?'' (select yes/no)
    \end{enumerate}
  }
\end{enumerate}

\end{document}